%% file: main.tex
\pdfoutput=1
\documentclass[tightenlines, twocolumn, notitlepage, nofootinbib, superscriptaddress]{revtex4}
\usepackage[dvipsnames]{xcolor}
\usepackage[intlimits]{amsmath}
\usepackage{amssymb}
\usepackage{bm}
\usepackage{bibunits}
\usepackage{changes}
\usepackage{enumerate}
\usepackage{enumitem}
\usepackage{floatrow}
\usepackage{hyperref}
\usepackage[symbol]{footmisc}
\usepackage{graphicx}
\usepackage{units}
\usepackage{xspace}
\usepackage{booktabs}
\usepackage{listings}
\usepackage{comment}
\usepackage{makecell}

\usepackage{dsfont}
\usepackage{multirow}

\usepackage[utf8]{inputenc}

\usepackage{floatrow}
\newfloatcommand{capbtabbox}{table}[][\FBwidth]
\usepackage{blindtext}

\graphicspath{{./}, {figures/}}

\newcommand{\gemini}{\textsc{Gemini}\xspace}
\newcommand{\phoenics}{\textsc{Phoenics}\xspace}

\newcommand{\gpflow}{\textsc{Gpflow}\xspace}

\newcommand{\olympus}{\textsc{Olympus}\xspace}
\newcommand{\gryffin}{\textsc{Gryffin}\xspace}

\makeatletter
\renewcommand*{\p@subsection}{\thesection.}
\makeatother

\begin{document}
	\title{\large{Gemini: Dynamic Bias Correction for Autonomous Experimentation\\ and Molecular Simulation}}

	\date{\today}
    
    \author{Riley J. Hickman}
     \email{riley.hickman@mail.utoronto.ca}
    \affiliation{Department of Chemistry, University of Toronto, Toronto, ON M5S 3H6 Canada}
    \affiliation{Department of Computer Science, University of Toronto, Toronto, ON M5S 2E4 Canada}
    
    \author{Florian H\"{a}se}
    \affiliation{Department of Chemistry, University of Toronto, Toronto, ON M5S 3H6 Canada}
    \affiliation{Department of Computer Science, University of Toronto, Toronto, ON M5S 2E4 Canada}
    \affiliation{Department of Chemistry and Chemical Biology,  Harvard University, Cambridge, MA 02138, USA}
    \affiliation{Vector Institute for Artificial Intelligence, Toronto, ON M5S 1M1, Canada}
    
    \author{Lo\"{i}c M. Roch}
    \affiliation{Department of Chemistry, University of Toronto, Toronto, ON M5S 3H6 Canada}
    \affiliation{Department of Computer Science, University of Toronto, Toronto, ON M5S 2E4 Canada}
    \affiliation{Vector Institute for Artificial Intelligence, Toronto, ON M5S 1M1, Canada}
    \affiliation{Atinary Technologies Inc, 1006 Lausanne, VD, Switzerland}
    
    \author{Al\'{a}n Aspuru-Guzik}
     \email{aspuru@utoronto.ca}
    \affiliation{Department of Chemistry, University of Toronto, Toronto, ON M5S 3H6 Canada}
    \affiliation{Department of Computer Science, University of Toronto, Toronto, ON M5S 2E4 Canada}
    \affiliation{Lebovic Fellow, Canadian Institute for Advanced Research (CIFAR), Toronto, ON, M5S 1M1, Canada}
    \affiliation{CIFAR Artificial Intelligence Research Chair, Vector Institute, Toronto, ON, M5S 1M1, Canada}
	
	\input{abstract}
	\maketitle

\begin{bibunit}[unsrt]
	
	\input{introduction}

    \input{formulating}

    \input{analytic}

    \input{perovskites}

    \input{oercat}
    
    \input{conclusion}

	\section*{Acknowledgments}
    The authors would like to thank Professor Roger Grosse, Shengyang Sun, Guodong Zhang  and Dr. Matteo Aldeghi for contribution to fruitful discussions. 
    R.J.H. gratefully acknowledges the Natural Sciences and Engineering Research Council of Canada (NSERC) for provision of the Postgraduate Scholarships-Doctoral Program (PGSD3-534584-2019). F.H. acknowledges financial support from the Herchel Smith Graduate Fellowship and the Jacques-Emile Dubois Student Dissertation Fellowship. A.A.G. would like to thank Dr.~Anders Fr{\o}seth for his support. All computations reported in this work are performed on the Niagara supercomputer at the SciNet HPC Consortium~\cite{niagara1, niagara2}. SciNet is funded by the Canada Foundation for Innovation, the Government of Ontario, Ontario Research Fund - Research Excellence, and by the University of Toronto.
    
    \section*{Code availability}
    
    The source code of \gemini will be available on the following GitHub repository: \href{https://github.com/aspuru-guzik-group/gemini}{https://github.com/aspuru-guzik-group/gemini} under an MIT license. To install, use \texttt{pip install matter-gemini} (\textit{Python} $\geq$ 3.6 required).
    
    \section*{Data availability}

    All original data produced in this work is available upon reasonable request.
    
	\phantomsection\addcontentsline{toc}{section}{\refname}\putbib[main]
\end{bibunit}
\clearpage
\newpage
\begin{bibunit}[unsrt]
    \renewcommand{\thesection}{S.\arabic{section}}
	\onecolumngrid
	\subsection*{Supplementary Information}
	\setcounter{section}{0}
	\setcounter{subsection}{0}
	\input{supplementary}

	\putbib[main]
	
\end{bibunit}
\end{document}

%% file: abstract.tex
\begin{abstract}
Bayesian optimization has emerged as a powerful strategy to accelerate scientific discovery by means of autonomous experimentation. However, expensive measurements are required to accurately estimate materials properties, and can quickly become a hindrance to exhaustive materials discovery campaigns. Here, we introduce \gemini: a data-driven model capable of using inexpensive measurements as proxies for expensive measurements by correcting systematic biases between property evaluation methods. We recommend using \gemini for regression tasks with sparse data and in an autonomous workflow setting where its predictions of expensive to evaluate objectives can be used to construct a more informative acquisition function, thus reducing the number of expensive evaluations an optimizer needs to achieve desired target values. In a regression setting, we showcase the ability of our method to make accurate predictions of DFT calculated bandgaps of hybrid organic-inorganic perovskite materials. We further demonstrate the benefits that \gemini provides to autonomous workflows by augmenting the Bayesian optimizer \phoenics to yeild a scalable optimization framework leveraging multiple sources of measurement. Finally, we simulate an autonomous materials discovery platform for optimizing the activity of electrocatalysts for the oxygen evolution reaction. Realizing autonomous workflows with \gemini, we show that the number of measurements of a composition space comprising expensive and rare metals needed to achieve a target overpotential is significantly reduced when measurements from a proxy composition system with less expensive metals are available. 
\end{abstract}

%% file: introduction.tex
\section{Introduction} \label{sec:introduction}

Arguably, due to its ubiquitous use in several areas of science, engineering and economics, optimization is one of the most important areas of applied numerical science. Optimization tasks in chemistry and materials science are particularly challenging given that they are typically multi-dimensional, non-convex and often involve expensive to evaluate objectives. Therefore, development of efficient algorithmic strategies capable of identifying optimal experiment choices has been an active area of research for decades~\cite{coley_autonomous_nodate,coley_autonomous_nodate-1}. These strategies are usually formulated as traversing an  objective function which in chemistry and materials science describes how a measurable property varies with respect to changing conditions or parameters. Optimization strategies are tasked with navigating the parameter space with the goal of identifying parameters that correspond to a desired property value, e.g., the task of finding optimal reaction conditions that leads to a high yield. In the chemistry laboratory, evaluations of a response surface can involve many experimental steps (e.g. multi-step synthesis and evaluation of product yield) and can be time or resource consuming. It follows that the use of all available measurements to propose only the most informative parameters is required to reduce the number of evaluations and ultimately accelerate the discovery process. Strategies such as random search~\cite{baba_convergence_1981, bergstra_random_nodate}, gradient descent~\cite{lucia_chemical_1990,boyd_convex_2004, bubeck_convex_2015}, grid searches~\cite{fisher_design_1937,box_statistics_2005,anderson_doe_2016} and evolutionary algorithms~\cite{goldberg_genetic_1989,srinivas_genetic_1994,koza_genetic_1994} have been deployed and have found success in chemistry, but are typically most useful when the objective can be evaluated rapidly and at a low cost.

Recently, there has been a surge in practical applications of machine learning (ML) in chemistry and materials science~\cite{butler_machine_2018}. Techniques from ML have allowed chemists to predict the outcome of chemical reactions~\cite{coley_prediction_2017, schwaller_found_2018} and their yield~\cite{ahneman_predicting_2018}, accelerate the discovery of clean energy materials~\cite{langner_beyond_2020},
 and improve the prediction of protein structure~\cite{noe_boltzmann_2019,senior_improved_2020}. ML-driven approaches to optimization have also gained popularity in recent years. For example, Zhou \textit{et al.} highlighted the advantages of deep reinforcement learning for the optimization of microdroplet reactions~\cite{zhou_optimizing_2017}. Bayesian optimization (BO) is an experiment planning strategy that uses a ML model (traditionally a Gaussian process) to construct a surrogate to the true response surface learned from \emph{all} available observations before deciding on the next experiment to perform~\cite{shahriari_taking_2016}. Recently, it has been suggested that integration of BO approaches into the chemistry laboratory could have critical implications for efficient synthesis of functional chemicals and materials~\cite{hase_phoenics:_2018,hase_gryffin_2020,shields_bayesian_2021}.

Autonomous experimentation describes another sector where ML enables a change in the approach to scientific discovery with the goal of augmenting cutting edge computational and experimental tools with statistical techniques in a closed-loop workflow~\cite{flores-leonar_materials_2020,correa-baena_accelerating_2018,stein_progress_2019,coley_autonomous_nodate,coley_autonomous_nodate-1,hase_next-generation_2019,dimitrov_autonomous_2019,gromski_how_2019}.
BO has been a popular choice to drive prototype implementations of autonomous science~\cite{gongora_bayesian_2020,burger_mobile_2020,langner_beyond_2020,macleod_self-driving_2020}, while various other strategies have also shown merit~\cite{nikolaev_autonomy_2016,bedard_reconfigurable_2018}. Researchers have also begun to benchmark experiment planning algorithms on a wide variety of synthetic response surfaces and datasets derived from chemistry and materials science~\cite{felton_summit_2020,hase_olympus_2020,suram_benchmarking_2020-1}.


Evaluation of response surfaces in chemistry and materials science are typically expensive. Budgeted experimentation can be severely limited by, for example, cost of expensive reagents, computing time, and opportunity costs. However, there are often more than one method to evaluate a property with varying levels of accuracy and expense. Researchers will entertain an \textit{expensive} method if it leads to \textit{more accurate} measurements. Conversely, \textit{less accurate} methods are tolerated in some cases if they are \textit{less expensive} but provide a proxy-measurement for an expensive one. We maintain that the exploitation of less expensive measurements as proxies for more expensive measurements using ML is a key element in the acceleration of scientific discovery by means of autonomous experimentation.

To provide an explicit example, chemists are often confronted with finding the multi-component solvent mixture which best dissolves a particular solute. The trial-and-error approach to determining the optimal solvent composition using only experimental tools could be time and resource intensive. Associated costs include procurement of enough material to conduct potentially many sequential iterations and time associated with carrying out the individual trials. Molecular dynamics simulations can provide an estimate of solubility at a reduced overall cost.
An ML-driven optimization workflow which exploits both computation and experiment potentially running at the same time (e.g. in a laboratory and on a workstation or computing cluster, respectively) could reduce the number of expensive laboratory measurements needed to achieve a target solubility.

The idea of leveraging proxy-learning approaches has been known to the chemistry community for several years. Ramakrishnan \textit{et al.} added ML corrections to computationally inexpensive baseline quantum chemistry methods to predict more accurate target properties in what is referred to as the $\Delta$-machine learning approach~\cite{ramakrishnan_big_2015}. Aspuru-Guzik and co-workers used Gaussian processes (GPs) to calibrate theoretical results to experimental ones while designing acceptor materials for single-junction organic solar cells~\cite{pyzer-knapp_bayesian_2016, lopez_design_2017}. Yet, in all of these examples the calibration step relied on a sizable dataset of expensive evaluations, which renders these approaches inapplicable to discovery tasks where such a dataset is yet to be collected. Furthermore, even with a dataset of appreciable size, building a calibration model is a rigid approach whose generalization can break down for new examples which are out-of-distribution of the calibration set. 
In recent years, researchers have reported GP-based multi-fidelity BO strategies for materials design and discovery~\cite{tran_smf-bo-2cogp_2020,tran_multi-fidelity_2020,herbol_cost-effective_2020}, which have innovated around the CoKriging method to predict the objective function. 

In this work, we endeavour to enable proxy-learning in an autonomous discovery setting. In order to do so, we require a tool which (i) does not rely on the existence of a sizable dataset of expensive evaluations, (ii) can iteratively suggest informative evaluations of the expensive surface, (iii) can leverage inexpensive evaluations, (iv) the computational demand of the strategy should be affordable and ideally not exceed the order of a few minutes runtime on a standard laptop, (v) is scalable to regimes with thousands of measurements. Herein we report \gemini: a tool which addresses the aforementioned criteria using a flexible, computationally efficient approach based on carefully regularized neural networks (NNs). \gemini can be used for regression tasks where multiple sources of data are available and expensive training data is scarce. Also, \gemini's predictions can be used to construct a more informative surrogate model in a BO framework which contains insight from an inexpensive response surface. We show that our method works best for response surfaces with a high degree of positive rank correlation that it is flexible enough to capture a wide variety of non-linear correlations between response surfaces. As such, \gemini is most useful in cases where inexpensive measurements closely follow the expected trends of more expensive measurements.

This work is organized as follows: Section~\ref{sec:formulating_gemini} formulates \gemini mathematically and compares it to related approaches. Section~\ref{sec:analytic_tests} asseses the performance of \gemini in a regression setting on a series of synthetic response surfaces. Section~\ref{sec:perovskite_regression} showcases the utility of \gemini for predicting bandgaps of hybrid organic-inorganic perovskite materials. Finally, in Section~\ref{subsec:gemini_phoenics_theory} we detail combination of \gemini with the Bayesian optimizer \phoenics~\cite{hase_phoenics:_2018} and in Section~\ref{subsec:optimization_results} show how this optimizer can significantly reduce the number of expensive measurements necessary to identify points in high-dimensional composition spaces that correspond to active electrocatalysts for the oxygen evolution reaction. 

%% file: formulating.tex
\section{Formulating Gemini}
\label{sec:formulating_gemini}

\subsection{Background and architecture formulation} \label{subsec:background_formulation}

\begin{figure*}[htb]%
    \centering
    \includegraphics[width=1.0\textwidth]{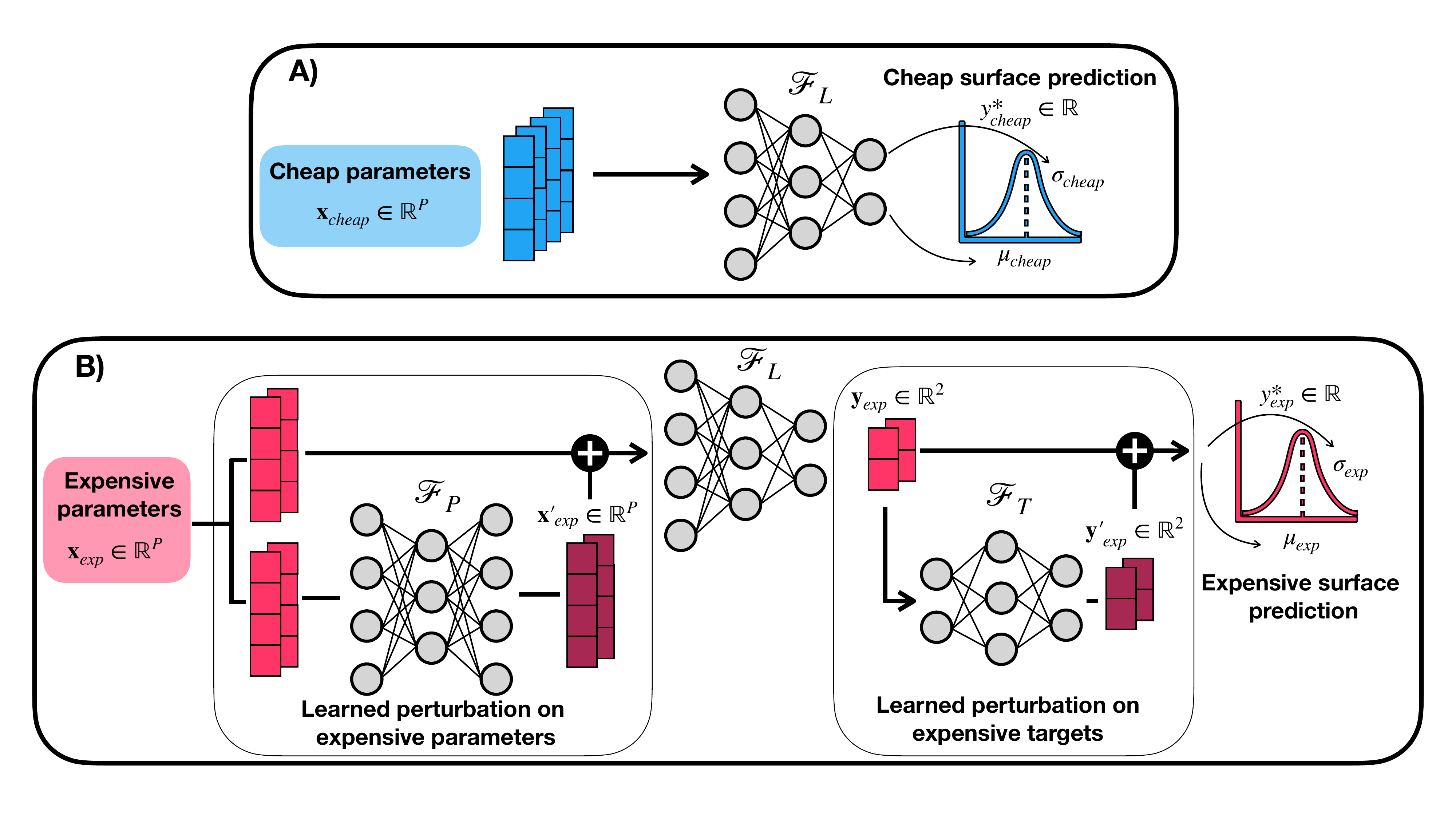}
    \caption{Outline of the \gemini model architecture for the prediction of a scalar-valued targets $y_{\text{exp}}$ and $y_{\text{cheap}}$ from a sets of parameters $\mathbf{x}_{\text{exp}} \in \mathbb{R}^P$ and $\mathbf{x}_{\text{cheap}} \in \mathbb{R}^P$. A) shows the forward pass of the model for the inexpensive parameters. $\mathbf{x}_{\text{cheap}} \in \mathbb{R}^P$ are passed directly through network $\mathcal{F}_L$ whose output is split to estimate the mean and variance of the target distribution. B) shows the forward pass of the model for expensive parameters. Here, $\mathbf{x}_{\text{exp}} \in \mathbb{R}^P$ are first passed though network $\mathcal{F}_P$ which learns a perturbation on the parameters outputting $\mathbf{x}_{\text{exp}} \in \mathbb{R}^P$ which is then added to the original expensive parameters and passed to $\mathcal{F}_L$. The output of $\mathcal{F}_L$ is then passed through $\mathcal{F}_T$ which intends to learn a perturbation on the expensive targets. The resulting output is then split to parameterize a distribution on the expensive targets. \gemini is trained in an end-to-end fashion, and weights and biases $\boldsymbol{\theta}$ for all three networks are updated via minimization of the distribution loss $\mathcal{L} = \mathcal{L} \left( \mu_{\text{exp}}, \sigma_{\text{exp}}, \mu_{\text{cheap}}, \sigma_{\text{cheap}}, \boldsymbol{\theta}\right)$ which contains additional regularization based on parameters $\boldsymbol{\theta}$ (see SI Sec.~\ref{subsec:architecture} for more information). Note that the sizes of the network illustrations do not necessarily reflect the hyperparameters chosen for \gemini (see SI Sec.~\ref{subsec:hyperparams}).}
    \label{fig:gemini_concept}%
\end{figure*}

We consider a scenario in which a researcher would like to determine the dependence of a scalar-valued target property, $y$, on a set of $p$ parameters $ \mathbf{x} = \{x_1,\ldots,x_p\}$. At the researcher's disposal there are two means of estimating $y$ for a given $\mathbf{x}$: an inexpensive, less accurate evaluator giving $ y_{\text{cheap}} = f_{\text{cheap}}(\mathbf{x})$, and a more expensive, more accurate evaluator giving $y_{\text{exp}} = f_{\text{exp}}(\mathbf{x})$. The researcher constructs datasets of observations $\mathcal{D}_{\text{cheap}} = \{(\mathbf{x}_{\text{cheap}, i}, y_{\text{cheap}, i}) \}_{i=1}^{N_c}$ and $\mathcal{D}_{\text{exp}} = \{(\mathbf{x}_{\text{exp}, i}, y_{\text{exp}, i}) \}_{i=1}^{N_e}$. We consider both regression type tasks (Sec.~\ref{sec:analytic_tests} and Sec.~\ref{sec:perovskite_regression}) where the goal is to fit a predictive model using pre-existing datasets, and autonomous discovery tasks (Sec.~\ref{sec:cat_oer_optimization}) where the datasets $\mathcal{D}_{\text{exp}}$ and $\mathcal{D}_{\text{cheap}}$ are constructed from scratch using sequential measurements with the goal of optimizing the property $y_{\text{exp}}$.

We are inspired by several previously reported approaches from the ML community. Analogously to chemistry, many objectives in ML research are costly to evaluate (i.e. optimizing model hyperparameters on a large dataset). Swersky \textit{et al.} proposed multi-task Bayesian optimization in which a small dataset is leveraged as a proxy for hyperparameter tuning on a larger set~\cite{swersky_multi-task_2013}. More recently, Astudillo and Frazier introduced Bayesian optimization of composite functions of the form $f(\mathbf{x}) = g(h(\mathbf{x}))$, where $h$ is an expensive to evaluate black-box function, and $g$ is an inexpensive to evaluate function. Transfer learning is a broadly defined field concerned with learning problems in which sufficient data is only available out-of-distribution from the task of interest, and a knowledge transfer model is applied~\cite{pan_survey_2010}. Few-shot learning can rapidly generalize learned prior knowledge to new tasks which contain only a few labelled examples~\cite{li_fei-fei_one-shot_2006,fink_object_2005}.
Bonilla \textit{et al.} investigate multi-task learning in the context of GPs~\cite{bonilla_multi-task_2008}. 
Multi-fidelity methods are intended to combine a small number of expensive \textit{high-fidelity} measurements with a greater number of cheaply-obtained \textit{low-fidelity} measurements~\cite{kennedy_predicting_2000,peherstorfer_survey_2018}. Recently, Cutajar \textit{et al.} reported deep GPs for multi-fidelity modeling, which allows for the capture of non-linear correlations between multi-fidelity data~\cite{cutajar_deep_2019}.

In this work, we seek to provide a tool which is able to use $\mathcal{D}_{\text{exp}}$ \textit{and} $\mathcal{D}_{\text{cheap}}$ to determine the set of parameters under which $f_{\text{exp}}$ yields an optimal response. Typically, optimization algorithms would only use $\mathcal{D}_{\text{exp}}$ to inform this decision as only $f_{\text{exp}}$ provides noisy ground truth objective measurements. However, with a clever choice of $f_{\text{cheap}}$ which may be motivated by an intuition-based belief that there exists some degree of statistical correlation between response surfaces of $f_{\text{exp}}$ and $f_{\text{cheap}}$, decision making using all available data may lower the overall expense required for the algorithm to find desired target values. We believe a practical realization the aforementioned tool would have the following characteristics:
\begin{enumerate} [label=(\roman*)]
    \item The tool should be able to correct for a wide class of non-linear biases between $f_{\text{exp}}$ and $f_{\text{cheap}}$ on-the-fly, i.e. in light of new observations of each surface at each optimization iteration. The tool should therefore be re-trained after every iteration and training times should not exceed the minute time scale for applicability in the chemistry and materials science sector.
    \item The tool should be robust to overfitting in the low-data regime, but also remain scalable to data abundant regimes as a general-purpose algorithm. Use cases may greatly vary in the number of iterations necessary to observe asymptotic behaviour of an optimizer. To ensure maximal applicability of our method, we wish for \gemini to be scalable from the low-data regime (10's of measurements) to potentially 1000's of measurements. This is motivated by the fact that cheap data is often available in abundance, and can be obtained in batches causing the total number of measurements (cheap + expensive) to increase rapidly.
    \item train using all available data, i.e. should not be mandatory that parameter points are measured with both $f_{\text{exp}}$ and $f_{\text{cheap}}$. This is a restrictive requirement of some calibration approaches where a model is trained to learn the difference between the two evaluators.
    \item The tool must be amenable to autonomous experimentation workflows, and therefore should satisfy standard requirements; e.g., not more than 20 dimensions, robust to noise, few expensive evaluations (usually no more than 100s).
\end{enumerate}
Consider the simple 1D case in which the $f_{\text{cheap}}$ and $f_{\text{exp}}$ are related in the following way,
\begin{align} \label{eq:bias_analytic_cheap}
    y_{\text{cheap}} = f_{\text{cheap}} \left( x\right). 
\end{align}
\begin{align} \label{eq:bias_analytic_exp}
    y_{\text{exp}} = f_{\text{exp}}\left(x\right) =  f_{\text{cheap}} \left[ x + f_p(x) \right] + f_t(x). 
\end{align} 
$f_p(x)$ represents a perturbation on the \textit{parameters}, while $f_t(x)$ is a perturbation on the \textit{targets}. Our model intends to learn the underlying function $f_{\text{cheap}}$ which maps paramerers to targets while simultaneously realizing the biases $f_p$ and $f_t$.

We use a NN based architecture which learns an abstract function mapping between parameters and targets while during the same training session learns systematic biases between the two sets of data. We choose NNs due to their favourable scaling relative to other probabilistic models such as GPs and their architectural flexibility which simplifies the definition of architectures which by construction can satisfy physical constraints such as positivity of a measured value. The abstract mapping is learned by $\mathcal{F}_{L}$ which is supplied data from both $\mathcal{D}_{\text{exp}}$ and $\mathcal{D}_{\text{cheap}}$. $\mathcal{F}_L$ estimates the mean and variance of the cheap and expensive target distributions through an auxiliary output unit, providing a means of estimating aleatoric uncertainty. Cheap parameters $\mathbf{x}_{\text{cheap}}$ are passed directly to $\mathcal{F}_L$ (Fig.~\ref{fig:gemini_concept} A), while a learned non-linear perturbation is added to $\mathbf{x}_{\text{exp}}$ through the use of the network $\mathcal{F}_P$ (Fig.~\ref{fig:gemini_concept} B). This step can be thought of as correcting for $``$shifts" between the parameters of each dataset (represented by $f_p$ in Eq.~\ref{eq:bias_analytic_exp}). Similar bias correction is performed in the target space, attempting to learn $f_t$ in Eq.~\ref{eq:bias_analytic_exp}. The raw output of $\mathcal{F}_L$ is  are perturbed using a representation learned by $\mathcal{F}_T$ to give the final expensive prediction, while the cheap prediction is the direct output of $\mathcal{F}_L$. The model is trained by minimizing the sum of the negative log-likelihood losses associated with the cheap and the expensive predictions. A conceptualization of the \gemini model is shown in Fig.~\ref{fig:gemini_concept} and a detailed description of the model architecture is given in SI. Sec.~\ref{subsec:architecture}. We provide a demonstration of \gemini's predictions on pairs of 1D synthetic response surfaces and compare them with those of NNs, GPs and random forests in SI Sec.~\ref{subsub:demonstration}.

%% file: analytic.tex
\section{Analytic tests} \label{sec:analytic_tests}

We envision \gemini as being most applicable in a global optimization setting and for closed-loop workflows for autonomous experimentation. It is important to first isolate the performance of \gemini on regression tasks. In this section, we assess the performance of \gemini in a regression setting using synthetic response surfaces.

\begin{figure}[!ht]%
    \centering
    \includegraphics[width=0.95\columnwidth]{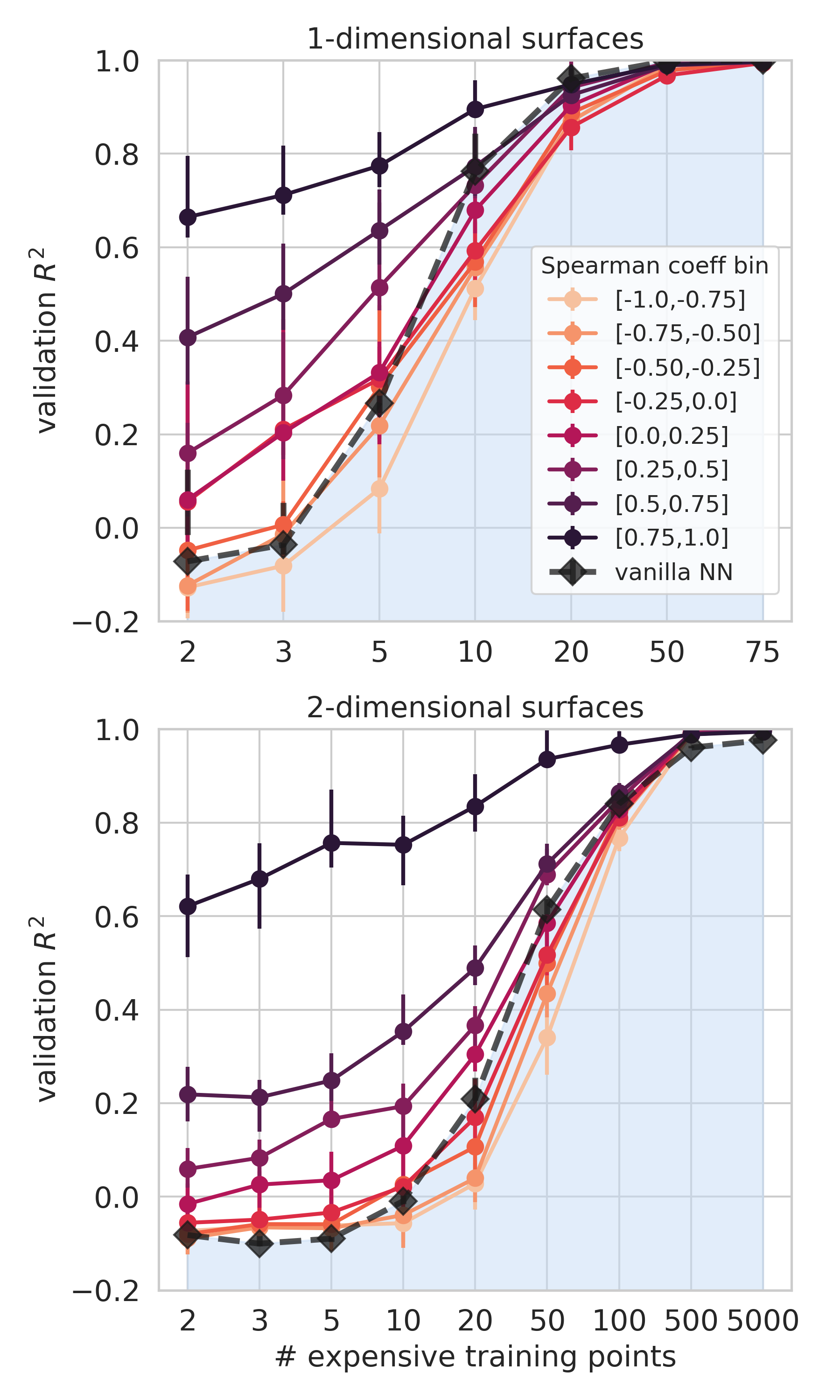}
    \caption{Performance of \gemini on 1D and 2D synthetic response surfaces with varying number of expensive training points. Each trace represents average performance over binned surfaces with similar Spearman coefficients, with 95\% confidence interval error bars. The black diamond trace represents the performance of a NN trained only on the expensive data (agnostic to the cheap data). The region shaded in light blue represents predictions that are less accurate than the vanilla NN. }
    \label{fig:sample_surface_pairs_1_2d}
\end{figure}

Synthetic surfaces when defined computationally can be used to thoroughly benchmark and explore the behavior of the suggested approach given that large scale evaluations are not costly in comparison to experimental evaluations. However, for an accurate assessment of the behavior and performance of \gemini on tasks related to chemistry and materials science, the synthetic surfaces should exhibit features generally expected from response surfaces encountered in these fields. Zhou \emph{et al.}~\cite{zhou_optimizing_2017} use Gaussian mixture models to generate a set of surfaces which displays properties found in chemistry and materials science: surfaces can have multiple local optima, are generally described by a manifold and are subject to noise, which we do not model explicitly. Here, we sample synthetic surfaces from a GP. A detailed overview of surface generation can be found in SI Sec. \ref{subsub:toy_generation_gp}. Surfaces are generated in pairs: one serving as the expensive surface, and one serving as the cheap surface and are populated with data points at random representing observations. 

For each pair of synthetic surfaces, a $``$vanilla" NN is trained on observations on the expensive surface only, while \gemini is trained on observations on both surfaces. Both methods make predictions on the same validation set which comprises the unobserved points on the expensive surface. The vanilla NN gives predictions which are agnostic to the cheap observations and serves as a regression baseline. We explore two axes with these tests.
\begin{enumerate} [label=(\roman*)]
    \item The relative performance of the two models as a function of the correlation of the cheap and expensive surfaces. Correlation is measured using Spearman's rank correlation coefficient, $r_s$, which measures statistical dependence between the rankings of two variables (full explanation in SI Sec. \ref{subsec:suppl_analytic_tests}).
    \item The relative performance of the two models when keeping the number of cheap observations fixed and increasing the number of expensive evaluations.
\end{enumerate}
The results of these tests for 1D and 2D GP derived surfaces are presented in Fig.~\ref{fig:sample_surface_pairs_1_2d}. In the 1D (2D) case, there are 75 (7500) cheap training points. The number of expensive training points is varied and models make a prediction on held-out validation sets of expensive points. The black diamond trace represents the vanilla NN baseline, while the coloured traces represent averaged performances of \gemini over pairs of surfaces categorized by their Spearman coefficient into 8 bins (see legend in Fig.~\ref{fig:sample_surface_pairs_1_2d}). In each bin, there are 20 pairs of surfaces. When few expensive training points are available, \gemini tends to outperform the vanilla NN for positively correlated surfaces, and displays the same asymptotic behaviour as the vanilla NN trace when the expensive training set size is increased. The margin of improvement using \gemini increases based on the increasing strength of positive correlation. Intuitively, choosing a cheap evaluator which correlates more strongly with the expensive evaluator should lead to better predictive accuracy with \gemini. For anti-correlated surfaces ($r_s\in[0.0, -1.0]$), \gemini performs as well or slightly worse than the NN. We note however that the drop in performance is not drastic. 

These tests demonstrate that \gemini is able to learn non-linear biases between response surfaces and significantly increase predictive performance in the low-data regime. Additional details about synthetic surface generation as well as higher-dimensional test results are reported in SI Sec. \ref{subsec:suppl_analytic_tests}). 
In the following section, \gemini is tested in a regression setting on real datasets taken from the chemistry and materials science literature. The \gemini hyperparameters used in the following tests are obtained by performing a hyperparameter grid-search using the average model performance over a set of 10 diverse synthetic response surface pairs as an objective. The motivation behind this approach is to find a set of hyperparameters that perform adequately across a wide range of data sizes and shapes. We hope that \gemini can then be used in an $``$out-of-the-box" fashion without the additional burden of finding application-specific hyperparameters. Information on the hyperparameter search can found in SI Sec.~\ref{subsec:hyperparams}. 


%% file: perovskites.tex
\section{Predicting bandgaps of hybrid organic-inorganic perovskites}
\label{sec:perovskite_regression}

\begin{figure}[htb]%
    \centering
    \includegraphics[width=0.99\columnwidth]{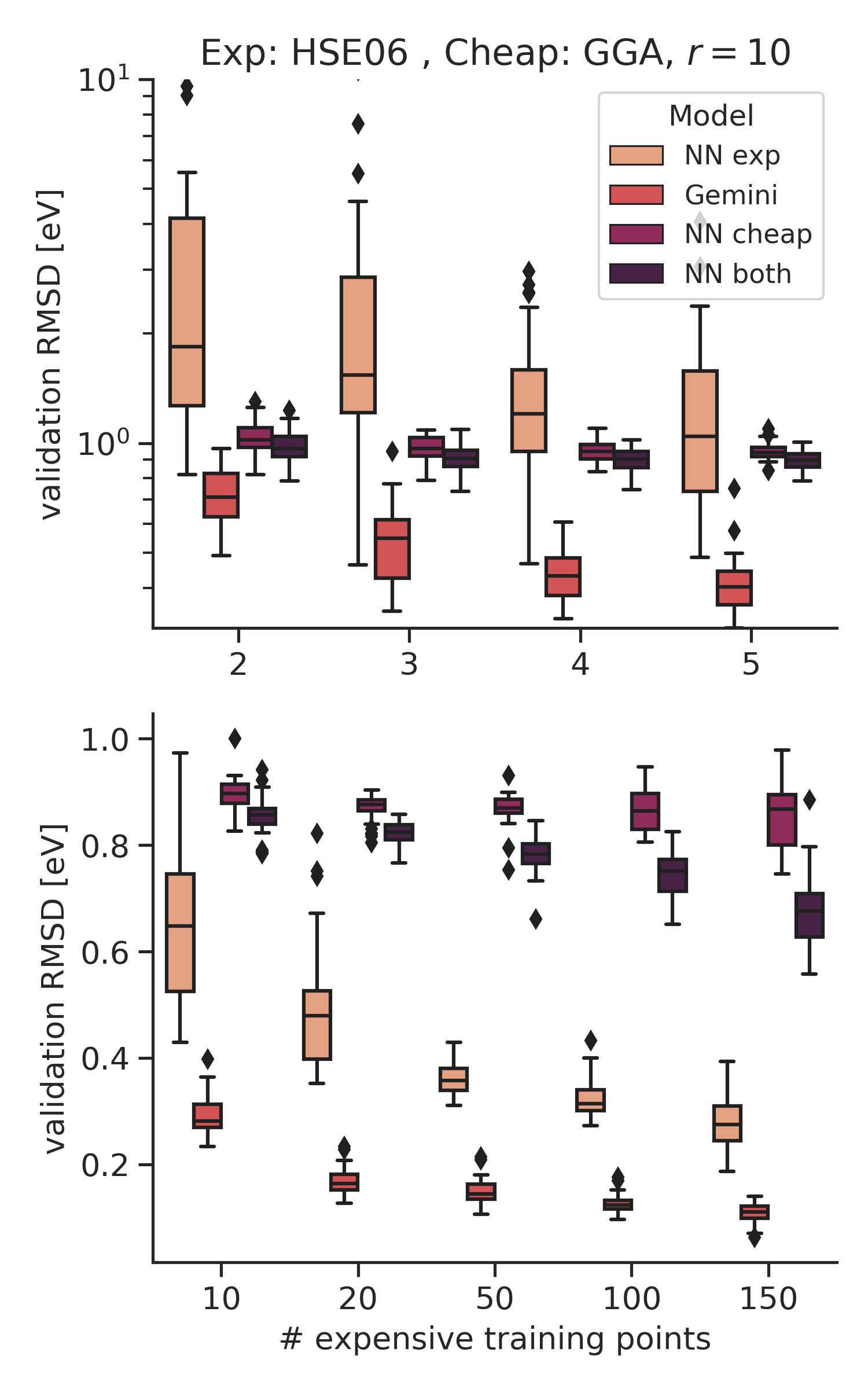}
    \caption{Learning curve distributions of validation prediction RMSDs for the HSE06 HOIP bandgaps. The number of cheap training points is $r=10$ times the number of expensive training points, with the number of cheap training points capped at the data set size (192 points). Reported prediction statistics encompass 40 randomly selected training/validation splits for each training set size. \textit{NN exp} indicates a network trained on the expensive data only, \textit{NN cheap} indicates a network trained on the cheap data only and making predictions on the expensive data, and \textit{NN both} indicates a network trained on both the expensive and cheap data. \gemini indicates the full architecture introduced in this work. \textit{NN exp}, \textit{NN cheap} \textit{NN both} constitute baseline predictive performance.}
    \label{fig:perovskite_regression}
\end{figure}

The efficiencies of perovskite solar cells have seen substantial improvements in the last decades, making them the one of the fastest advancing light-harvesting technologies~\cite{manser_intriguing_2016}. Hybrid organic-inorganic perovskites (HOIPs) are typically comprised of organic anions within lead halide coordination matrices~\cite{jeon_compositional_2015,nie_solar_2015,yang_analyzing_2019}. The inherent toxicity of lead motivates the exploration of new environmentally friendly HOIPs and is an active area of research~\cite{im_identifying_2019,sun_accelerated_2019,wu_metal-free_2020}. There are a vast number of perovskite designs accessible through chemical modification, however, accurate simulation of properties, device fabrication, and testing are prohibitively costly. In order to enable the accelerated identification of promising HOIPs via autonomous experimentation platforms it is crucial that accurate predictions of HOIP properties are available given sparse high-accuracy data.

A recently reported dataset of HOIPs contains all 192 combinations of materials assembled from 16 organic anions, 3 group-IV cations and 4 halide anions. The bandgap, $E_g$, is perhaps the most desired property of HOIP materials~\cite{kim_hybrid_2017}. The dataset reports $E_g$ for each HOIP material obtained using DFT calculations using both the generalized gradient approximation (GGA) and hybrid functionals. Calculated bandgaps using the Heyd–Scuseria–Ernzerhof (HSE06) exchange-correlation functional are expected to be close to experimentally determined bandgaps, but are computationally restrictive. The GGA level of theory is computationally feasible, but has been shown to underestimate $E_g$ by 30\%~\cite{perdew_density_1985}. We omit a detailed comparison of exchange-correlation approximations and roughly consider the HSE06 level of theory to be one order of magnitude more computationally expensive than the GGA level.

This example of the applicability of \gemini focuses on the prediction of expensive $E_g^{\text{HSE06}}$ values using both cheap (GAA) and expensive (HSE06) data and showcases how \gemini can generate accurate predictions in a setting where training data is available at varying levels of computational expense. Perovskite candidates are featurized using geometric and electronic descriptors of their components. Namely, the electron affinity, ionization energy, total mass and electronegativity of the inorganic cations and anions are used, as well as the HOMO-LUMO energies, dipole moment, atomization energy, radius of gyration and total mass of the organic anions. Cumulatively, each material is described by a 14 element vector. These descriptors were found to be most informative for optimization of the bandgap over the same dataset in Ref.~\cite{hase_gryffin_2020}. Individual descriptor values and further explanation are given in SI Sec.~\ref{subsec:perovskite_details}.

We conduct a series of regression tests where the predictive performance on the expensive dataset of 4 methods are compared: i) a NN trained on the expensive data only (\textit{NN exp}), ii) \gemini trained on cheap \emph{and} expensive data, iii) a NN trained on the cheap data \textit{NN cheap}, and iv) a NN trained on both the cheap\emph{and} expensive data (\textit{NN both}). All models make predictions on a validation set of expensive measurements. Methods iii) and iv) provide simple transfer learning baselines. For each training set size, 40 training sets are randomly sampled from the entire dataset and the remainder of the points are used for prediction. Prediction statistics are shown in Fig.~\ref{fig:perovskite_regression}. In line with our estimation of density functional expense, the number of cheap training data points is 10 times the number of expensive training points. e.g. \textit{NN exp} is trained using 5 expensive points only, \gemini and \textit{NN both} are trained using 5 expensive points and an additional 50 cheap points, while \textit{NN cheap} is trained using the 50 cheap points only. For the cases where 10 times the number of expensive points exceeds the size of the dataset (i.e. 20 expensive points and up), the number of cheap points available to \gemini, \textit{NN both}, and \textit{NN cheap} is the entire dataset (192 points).

Fig.~\ref{fig:perovskite_regression} shows learning curve prediction statistics over the 40 random training set samples for each training set size. \gemini makes more accurate predictions than the other 3 models when expensive data is scarce (2-10 expensive points), when there is moderate expensive data (10-100 expensive points) and when expensive data is plentiful (150 expensive points). We find that as the amount of expensive data is increased, the difference between the performance of \gemini and NN exp decreases, however, even with 150 expensive points ($\sim$78\% of the dataset) the additional 42 cheap data points are used by \gemini to further reduce the validation RMSD. We observe that the predictions of NN both and NN cheap approaches do not become significantly more accurate with increasing training data. This is due to a systematic underestimation of the bandgap using GAA and the inability of these methods to correct this simple bias (see SI Sec.~\ref{subsec:perovskite_details} for more details on the dataset). This example demonstrates the ability of \gemini to use data from different sources accurately predict expensive to evaluate properties. We stress that the applicability of \gemini is not constrained to purely computational measurements. From the point of view of the algorithm, experimental determination of HOIP bandgap could instead be used as the expensive evaluator.

%% file: oercat.tex
\section{Optimizing electrocatalytic activity of high-dimensional composition spaces for the oxygen evolution reaction}
\label{sec:cat_oer_optimization}

\subsection{Combining \gemini with a Bayesian optimization strategy}
\label{subsec:gemini_phoenics_theory}

\begin{figure}[htb]%
    \centering
    \includegraphics[width=0.99\columnwidth]{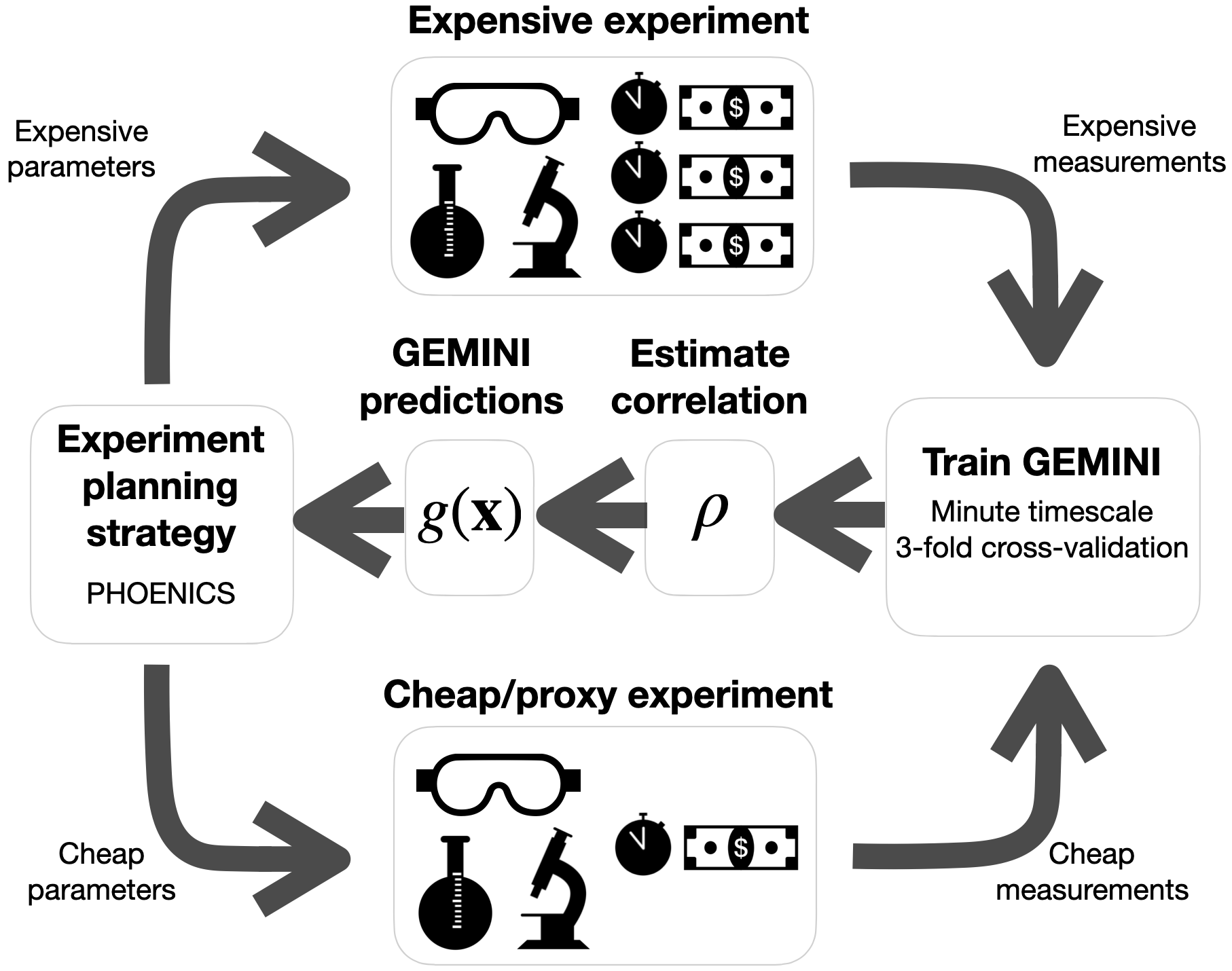}
    \caption{General schematic showing how \gemini can be integrated into a closed-loop workflow which contains multiple sources of measurement. An experiment planning strategy (\phoenics in this work) recommends parameters to be measured by the cheap and expensive evaluators. In this figure, the expensive experiment is depicted as carrying a greater temporal and  monetary cost. The resulting measurements are used to train \gemini, after which the correlation ($\rho$ in Eq.~\ref{eq:gemini_acquisition_function}) between \gemini's predictions $g(\mathbf{x})$ and the true expensive measurements is estimated using a cross-validation approach (see SI Sec.~\ref{subsec:details_opt_cat_oer}). In this work, $\alpha_g(\mathbf{x})$ is then computed and used to select parameters to be evaluated at the subsequent iteration of the loop.}
    \label{fig:closed_loop_schematic}
\end{figure}

A global optimization algorithm should be able to identify sets of parameters for which a desired objective value is achieved. As was shown in Sec.~\ref{sec:analytic_tests} and Sec.~\ref{sec:perovskite_regression}, \gemini can make accurate predictions of expensive to evaluate response surfaces when supplemented by a correlated proxy-data source. By re-training \gemini in light of new observations and incorporating its predictions into the acquisition function of a Bayesian optimizer, \gemini can be effectively utilized in a closed-loop workflow. We report integration of \gemini with the Bayesian optimizer \phoenics~\cite{hase_phoenics:_2018}. \phoenics is a probabilistic global optimization algorithm that combines ideas from Bayesian kernel density estimation and Bayesian optimization and is already designed to be useful for optimization problems in which evaluations are time consuming or resource intensive. \phoenics uses kernel densities estimated by a Bayesian neural network (BNN) and scaled by the observed objective function values to construct a surrogate model.

To accommodate \gemini, we propose a modification to the acquisition function of \phoenics in which one term is added in the numerator and denominator.
\begin{align} \label{eq:gemini_acquisition_function}
        \alpha_g(\mathbf{x}) = \frac{\sum_{k=1}^{n} f_k p_k(\mathbf{x}) + \lambda p_{\text{uniform}}(\mathbf{x}) + \rho g(\mathbf{x})}{\sum_{k=1}^{n} p_k(\mathbf{x}) + p_{\text{uniform}}(\mathbf{x}) + \mathds{1}}\,.
\end{align}
$g(\mathbf{x})$ is \gemini's prediction of the objective function at the observed points. $\rho$ is Pearson's correlation coefficient between $g(\mathbf{x})$ and the true observed values $f_k$. When $g(\mathbf{x})$ becomes increasingly correlated or anti-correlated with the $f_k$ (corresponding to a $\rho$ approaching 1 and -1 respectively), the more pronounced $g(\mathbf{x})$ becomes in Eq.~\ref{eq:gemini_acquisition_function}. When $g(\mathbf{x})$ is completely uncorrelated with the true function values ($\rho = 0$), Eq.~\ref{eq:gemini_acquisition_function} becomes agnostic to the prediction of \gemini. The qualitative impact of both $\lambda$ and $\rho$ on the shape of $\alpha_g(\mathbf{x})$ are illustrated in SI Sec.~\ref{subsec:lambda_rho_acq}.

Augmentation of \phoenics with \gemini provides a scalable optimization algorithm compatible with multiple sources of measurement. Successfully combining this optimizer with such sources (e.g. high-performance computing, robotic platforms, automated laboratory equipment, etc.) would fulfill our goal of enabling proxy-learning in an autonomous discovery setting. Fig.~\ref{fig:closed_loop_schematic} shows a succinct depiction of how \gemini can be integrated into a closed-loop scientific discovery workflow comprising multiple sources of property measurement. An experiment planning strategy, such as \phoenics, recommends parameter points to be measured by the cheap and the expensive experiment. The expensive experiment may incur a significantly greater temporal or monetary cost per measurement (Fig.~\ref{fig:closed_loop_schematic}). The resulting property measurements are used to train \gemini, which in turn provides an estimation of the correlation between its predictions $g(\mathbf{x})$ and the true property values ($\rho$ in Eq.~\ref{eq:gemini_acquisition_function}). $\alpha_g(\mathbf{x})$ is then computed and is used by the experiment planner to select parameter points to be evaluated in the next iteration (see see SI Sec.~\ref{subsec:details_opt_cat_oer} for additional details). In Sec.~\ref{subsec:optimization_results}, we simulate such an autonomous discovery platform including \phoenics and \gemini inspired by a recently reported dataset of electrocatalyst activities for the oxygen evolution reaction.

\subsection{Results of optimization experiments}
\label{subsec:optimization_results}


The discovery of active electrocatalysts for the oxygen evolution reaction (OER) is an important challenge in energy materials science~\cite{back_toward_2019,song_review_2020}. The rise of scientific instrument automation has enabled catalyst discovery to benefit from previously unparalleled amounts of data collected using high-throughput experimentation. Recently, OER activity and catalytic stability have been measured through systematic exploration of high-dimensional chemical spaces~\cite{soedarmadji_tracking_2019,stein_functional_2019}. The resulting four datasets each contain a discrete library of 2121 catalysts, comprising all unary, binary, ternary and quaternary compositions from unique 6 element sets with 10 at\% intervals. These datasets have already been used to benchmark sequential learning strategies~\cite{suram_benchmarking_2020}. 

The four composition systems can be ranked with respect to expense in several ways, including monetary cost of the elements in the sets, the abundance of the elements in the earths crust, or the toxicity of the elements. We only consider monetary cost in this work. Table~\ref{tab:cat_oer_expense} describes the composition systems and the cost associated with procuring an equal mass of all the elements. Considering these costs, pairs of composition systems can be classified as \emph{cheap} or \emph{expensive}. For instance, evaluations of the C compositions are on average roughly 6.4 times as expensive as evaluations of the A compositions due to Tantalum being a rare element in high demand. We designate the C composition measurements as expensive and the A composition measurements as cheap. 

\begin{table}[htb] \label{tab:cat_oer_expense}
    \begin{center}
    \begin{tabular}{ | c | c | c |  p{5cm} |} 
    \hline
    \multirow{2}{*}{} \textbf{Label} & \textbf{Composition} & \textbf{Price}  \\ 
                                     &  \textbf{system} & \textbf{[USD/g]}  \\ \hline
     A  & Mn-Fe-Co-Ni-La-Ce &  0.06  \\ \hline 
     B  & Mn-Fe-Co-Ni-Cu-Ta &  0.37  \\ \hline
     C  & Mn-Fe-Co-Cu-Sn-Ta &  0.37  \\ \hline
     D  & Ca-Mn-Co-Ni-Sn-Sb &  0.08  \\ 
    \hline
    \end{tabular} 
    \caption{Labels, elemental composition, and estimated monetary cost of each system reported in Ref~\cite{suram_benchmarking_2020}. Prices of individual elements are taken from 3 different sources.~\cite{bgr_metals,smm_metals,ise_metals} The price column contains the amount in USD needed to purchase 1g of the composition assuming an equal mass of each element.}
    \end{center}
\end{table}

\begin{figure}[!ht]%
    \begin{center}
    \includegraphics[width=0.99\columnwidth]{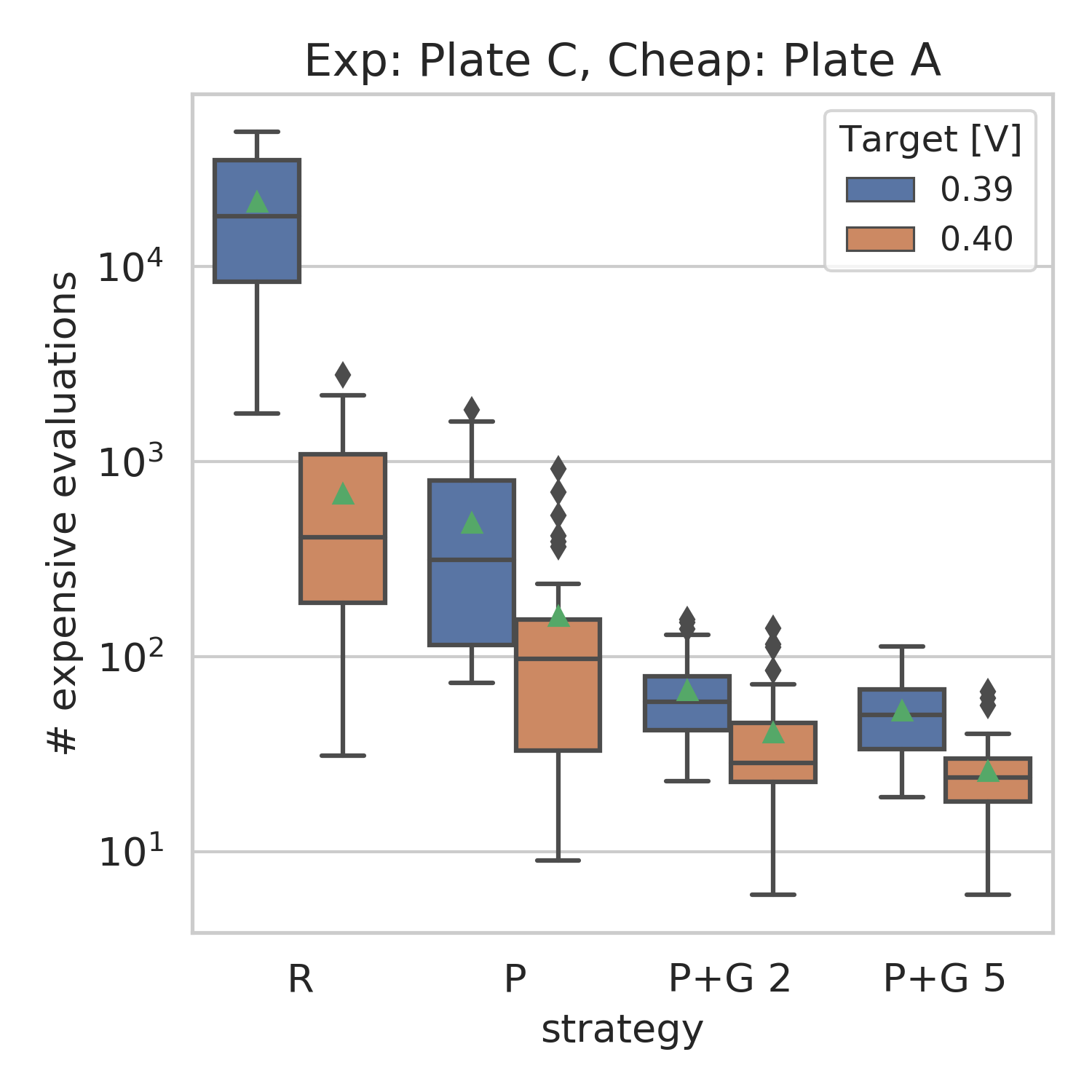}
    \end{center}
    \caption{ Comparison of the number of expensive evaluations taken by optimization strategies to achieve two targeted overpotentials (0.40 V and 0.39 V) on the emulated plate C dataset. The emulated plate C (A) datasets constitute the expensive (cheap) surfaces respectively. The strategies are abbreviated as follows: R: random search (no cheap surface evaluations), P: Phoenics (no cheap surface evalautions), P+G $r$: Phoenics combined with Gemini, where $r$ is the number of cheap surface evaluations taken per expensive evaluation. Box-and-whisker plots are generated using 36 independent runs for each strategy. Green triangles represent average expensive evaluations needed to reach the targets.}
    \label{fig:cat_oer_opt}
\end{figure}
\begin{table}[!ht] \label{tab:cat_oer_opt_res}
\begin{tabular}{|c|c|c|}
\hline
       & \multicolumn{2}{c|}{\textbf{Target overpotential}} \\ \hline
\textbf{Strategy} & \textbf{0.39 V}      & \textbf{0.40 V}     \\ \hline
\multirow{2}{*}{Random sampling }   &  21157 $\pm$ 2456        & 618 $\pm$ 106          \\
                                    & N/A & N/A \\ \hline

\multirow{2}{*}{\phoenics only}  &  486 $\pm$ 78          & 161 $\pm$ 33          \\ 
                                 & ($1.7\times10^{-8}$) & ($8.6\times10^{-5}$)\\ \hline

\multirow{2}{*}{\phoenics + \gemini}  &  67 $\pm$ 6   & 41 $\pm$ 5     \\ 
                                    $r=2$ & ($6.7\times10^{-7}$)  & ($2.2\times10^{-4}$)  \\ \hline
\multirow{2}{*}{\phoenics + \gemini}  &  \textbf{53 $\pm$ 4}   &  \textbf{26 $\pm$ 2} \\ 
                            $r=5$  & ($1.8\times10^{-2}$)  &  ($6.9\times10^{-3}$) \\ \hline
\end{tabular}
\caption{Mean expensive surface evaluations taken for strategies to find emulated C compositions corresponding to target overpotentials. Averages are taken over 36 independently seeded runs for each strategy and standard errors on the mean are reported. The strategy needing the fewest expensive surface evaluations to achieve each target is bolded. p-values from a paired difference Wilcoxon sign test are shown in parentheses between each method and the method directly above it in the table, e.g. P+G 5 compared to P+G 2, P+G 2 compared to P, and so on. Full box-and-whisker plots for these runs are shown in  Fig.~\ref{fig:cat_oer_opt}  }
\end{table}
To showcase the performance of \gemini in a continuous-valued parameter optimization framework, we are inspired to construct a probabilistic emulator (virtual robot~\cite{hase_chimera:_2018,hase_olympus_2020}) using a BNN to reproduce the OER catalysts datasets. This approach allows for the extension of a discrete library of measurements to a continuous function that can be queried for virtual measurements at any point in parameter space. It is important to note that we construct an emulator to reproduce the entire standard 6-simplex even though discrete values from Ref.~\cite{suram_benchmarking_2020} only include up to quaternary compositions. The reader is referred to SI Sec.~\ref{subsec:details_opt_cat_oer} for further details on the BNN emulator and justification for this approach. The goal is to minimize the overpotential of the emulated C composition system while using the emulated A composition system as proxy measurements. Results are displayed in Fig.~\ref{fig:cat_oer_opt}. We compare four distinct strategies. R: Random sampling of the expensive surface (no cheap surface evaluations), P: \phoenics optimizing the expensive surface (no cheap surface evaluations), P+G 2: \phoenics + \gemini with 2 cheap surface evaluations per expensive surface evaluation, and P+G 5: \phoenics + \gemini with 5 cheap surface evaluations per expensive surface evaluation. All strategies using \phoenics use two sampling strategies ($\lambda=1,-1$) and parameters for cheap surface evaluations are chosen randomly. More involved cheap surface sampling strategies such as active learning~\cite{settles_active_2012} which attempt to resolve the entire cheap surface with as few evaluations as possible should be considered in subsequent work. We conduct the search/optimization sequentially until a target expensive surface overpotential of 0.39 V is achieved. Fig.~\ref{fig:cat_oer_opt} shows the number of expensive surface evaluations taken for each strategy to achieve the target. We also show results for a larger overpotential target of 0.40 V. Box-and-whisker plots show statistics over 36 independent runs, and mean expensive surface evaluations for each strategy are summarized in Table \ref{tab:cat_oer_opt_res}. \phoenics alone is able to drastically reduce the number of evaluations from random sampling for both targets showing the utility of BO for this problem, although more than 100 evaluations are needed (typically large for BO applications). \phoenics augmented with \gemini exploiting 2 cheap evaluations per expensive evaluation takes roughly 4 (7) times fewer expensive evaluations to achieve the 0.40 V (0.39 V) target than \phoenics only. To achieve the 0.39 V overpotential target, P+G 2 needs on average only 67 expensive evaluations compared to 486 with \phoenics only. To achieve the 0.40 V target P+G 2 needs only 41 expensive evaluations compared to  161 with \phoenics only.  Exploiting 5 cheap evaluations per expensive evaluation needs about 6 (9) times fewer expensive evaluations to achieve the 0.40 V (0.39 V) target than \phoenics only and further reduces the average number of expensive evaluations needed to achieve each target to 53 and 26, respectively. This demonstrates that increasing the number of cheap evaluations available per expensive evaluation can further reduce the number of expensive measurements needed for the \phoenics + \gemini optimizer to reach a target value. 

%% file: conclusion.tex
\section{Conclusions}
\label{sec:conclusion}


We have reported \gemini, a machine learning tool which can leverage inexpensive measurements as a proxy for expensive measurements in an autonomous experimentation setting. The implementation of \gemini is flexible enough to correct for a wide class of non-linear biases between response surfaces, and gives the most accurate predictions when there is a strong, positive rank correlation between the response surfaces. We demonstrate that our method can give accurate predictions of perovskite bandgap energies at the HSE06 level when given proxy measurements at the GGA level of theory. \gemini can be combined with the Bayesian optimizer \phoenics to yield a scalable optimization strategy compatible with multiple sources of measurement. Using our method, we simulate optimization of overpotential of oxygen evolution reaction electrocatalysts over a 6-element composition system using overpotential measurements from a separate, cheaper composition system as proxies. We show that \phoenics augmented with \gemini can greatly reduce the number of expensive measurements needed achieve a target overpotetnial value compared to traditional optimization approaches. Thus, we maintain that \gemini contributes to the advancement of optimization strategies for autonomous experimentation as a tool that can make use of inexpensive data. \gemini is provided as an open source \textit{Python} package available on GitHub (\href{https://github.com/aspuru-guzik-group/gemini}{https://github.com/aspuru-guzik-group/gemini}) under an MIT license. To install, use \texttt{pip install matter-gemini} (\textit{Python} $\geq$ 3.6 required).

In future studies, we anticipate \gemini to be added to categorical BO frameworks. Categorical BO is essential for the design of functional molecules and advanced materials and could benefit from the proxy-learning approach detailed in this work. In particular, we have recently reported \gryffin, which extends the idea of \phoenics to the selection of categorical variables (e.g., catalysts, solvents). Our method could, in theory, be added to other experiment planning strategies that are potentially more suited for a particular application than BO. \gemini could also be applied to not only learn biases between different methods of evaluating the same property, but between \emph{different properties} entirely with varying difficulty associated with their measurement. For instance, Bayesian ML has been used to correlate electronic absorption spectra of PEDOT:PSS nanoaggregates (cheap) with the strength of intermolecular electronic couplings in organic conducting and semiconductor materials (expensive)~\cite{roch_absorption_2020}. Similar applications could broaden the applicability of \gemini.

%% file: supplementary.tex

\section{Analytic tests} \label{subsec:suppl_analytic_tests}

\subsection{Demonstration of \gemini's predictions and comparison to other models} \label{subsub:demonstration}

\begin{figure}[!ht]%
    \begin{center}
    \includegraphics[width=\textwidth]{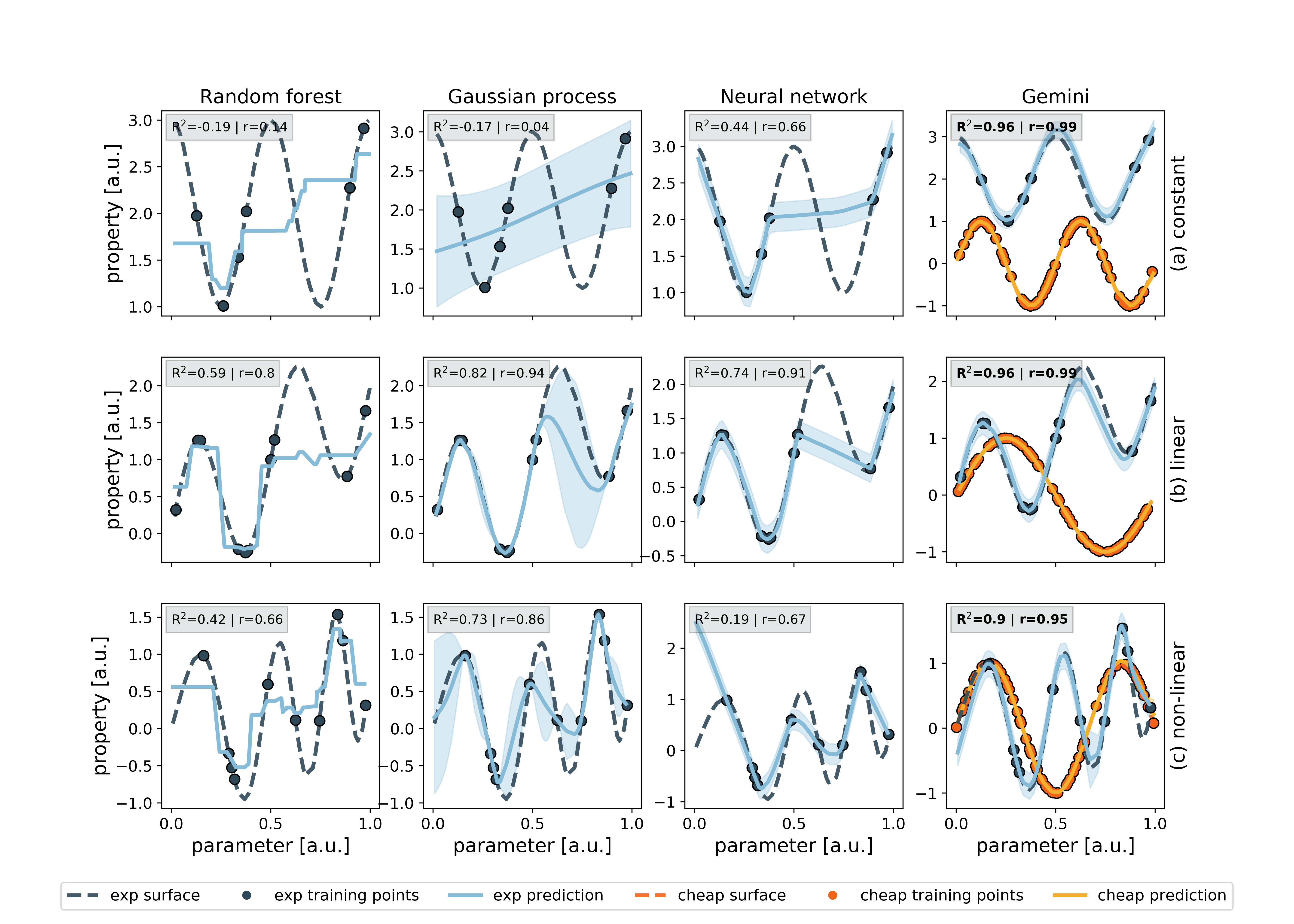}
    \end{center}
    \caption{Comparison of predictions of three baseline models and \gemini on 1D trigonometric surfaces which are related by perturbative biases on the parameters and targets (surfaces and biases are summarized in Table~\ref{tab:toy_functions_eq}). a) constant bias, b) linear bias, c) non-linear bias. Baseline models consist of a random forest, a GP, and a neural network and are trained on the expensive training points only. \gemini is trained on both expensive and cheap measurements. Shaded regions represent +/- the predicted variance of the respective model. $R^2$ is the coefficient of determination and $r$ is Pearson's correlation coefficient. Both statistics are reported on the expensive test set. The best performing models for each row are shown in bold.
    }
    \label{fig:gemini_trig_tests}%
\end{figure}

As a simple example, we examine the performance of \gemini using 3 pairs of 1D trigonometric functions as the cheap and expensive surfaces. Each pair represents a different class of bias functions $f_p$ and $f_t$: (a) constant bias, (b) linear bias, and (c) non-linear bias. Analytic forms of these functions are summarized in Table~\ref{tab:toy_functions_eq}. Fig. \ref{fig:gemini_trig_tests} illustrates the results of the 1D regression examples. All surfaces are plotted on the domain $[0,1]$. Three different models are trained on the expensive measurements only (blue dots), and predict the remaining test points. These models are a random forest (as implemented in Scikit-learn~\cite{sklearn_2011}), a GP (Matérn 3/2 kernel, as implemented in GPFlow~\cite{matthews_gpflow_2017,van_der_wilk_framework_2020}), and a fully-connected neural network. In the right-most column, \gemini is trained on both the expensive and cheap measurements (orange dots) and predictions are shown along with uncertainty in blue (orange) for the expensive (cheap) surfaces. Test set prediction accuracy statistics on the expensive surface are given for the baseline predictive models and for \gemini in their respective columns. For each class of bias functions, we find that \gemini's prediction and the true expensive surface have near perfect linear correlation (Pearson's correlation coefficient, $\rho \approx 1$), which is not always the case for the baseline predictive models. The coefficient of determination ($R^{2}$), a more general non-linear correlation statistic is also significantly higher for the predictions of \gemini than the predictions of the baseline models. This simple example indicates that using \gemini can give accurate predictions of expensive to evaluate response surfaces when cheap measurements from a correlated proxy-surface are plentiful. \gemini appears to be able to learn constant, linear, and non-linear classes of biases between the surfaces even when expensive measurements are sparse (5-10 measurements).

\begin{center}
\begin{table}[!htb]
\label{tab:toy_functions_eq}
\begin{tabular}{|c|c|c|c|c|}
\hline
\textbf{Type}       & \begin{tabular}[c]{@{}l@{}}\textbf{Cheap}\\ \textbf{surface}\end{tabular} & \begin{tabular}[c]{@{}l@{}}\textbf{Expensive}\\ \textbf{surface}\end{tabular} & $f_p(x)$ & $f_t(x)$ \\ \hline
constant (a)   &  $\cos4\pi x+2$  & $\sin4\pi x$  & $3/8$  &  $-2$   \\ \hline
linear (b)    &  $\sin2\pi x$ & $\sin4\pi x +2x$  & $x$ & $2x$  \\ \hline
non-linear (c) & $\sin 3\pi x$ & $\sin3\pi \left(x+x^2\right) +x^3$ & $x^2$ & $x^3$ \\ \hline

\end{tabular}
\caption{1D Trigonometric functions representing toy surfaces used to test the performance of \gemini and the bias functions between them. Bracketed letters in the Type column indicate the row in Fig.~\ref{fig:gemini_trig_tests} containing a visualization of the example.}
\end{table} 
\end{center}

\subsection{Comparison of synthetic response surfaces} \label{subsub:comparsion_of_surfaces}

\begin{enumerate}

 \item \textbf{Spearman's Rank Correlation Coefficient}: Spearman's Rank correlation coefficient, $r_s$, is a nonparametric measure of statistical dependence between the rankings of two variables. The raw vectors of cheap and expensive observations $\mathbf{y}_{\text{cheap}}$ and $\mathbf{y}_{\text{exp}}$ are transformed to the rank vectors $\mathbf{r}_{\text{cheap}}$ and $\mathbf{r}_{\text{exp}}$ 
 which contain integer ranks for each value in ascending order. $r_s$ is then computed as Pearson's correlation coefficient between these rank vectors.
 \begin{align} \label{eq:spearman}
    r_s = \rho_{\mathbf{r}_{\text{cheap}}, \mathbf{r}_{\text{exp}}} = \frac{\text{cov}\left( \mathbf{r}_{\text{cheap}}, \mathbf{r}_{\text{exp}}\right)}{\sigma\left( \mathbf{r}_{\text{cheap}}\right) \sigma\left(\mathbf{r}_{\text{exp}}\right)}\,.
 \end{align}
 The numerator denotes the covariance of the rank variables, and the denominator is the product of the rank variables standard deviations. A positive (negative) $r_s$ indicates that $\mathbf{y}_{\text{cheap}}$ increases (decreases) monotonically with $\mathbf{y}_{\text{exp}}$. 

  \item \textbf{Ratio of expensive to cheap observations}: In the analytic tests we keep the number of cheap training points fixed while varying the number of expensive training points.

\end{enumerate}

\subsection{Generation of synthetic response surfaces using Gaussian processes} \label{subsub:toy_generation_gp}

Synthetic benchmark functions derived computationally provide an efficient means of thoroughly benchmarking the performance of \gemini. Pairs of synthetic response surfaces are generated by sampling from a Gaussian process (GP) prior. GPs are generalizations of probability distributions over variables to functions~\cite{rasmussen_gaussian_2006}, defined by a mean function $\mu(\mathbf{x})$ and a positive definite covariance function $k(\mathbf{x}, \mathbf{x}')$, with $(\mathbf{x}, \mathbf{x}')$ representing all possible pairs of input domain points. 
\begin{align} \label{eq:gaussian_process}
    f(\mathbf{x}) \sim \mathcal{GP} \left(\mu(\mathbf{x}), k(\mathbf{x}, \mathbf{x}') \right).
\end{align}
In order to model the variability of the GP random variables, we must define a covariance function. The radial basis function (RBF) kernel, which is a popular choice in kernelized learning algorithms. The RBF kernel for two sample points $\mathbf{x}$, $\mathbf{x'} \in \mathbb{R}^k$ is defined as
\begin{align} \label{eq:rbf_kernel}
    k(\mathbf{x}, \mathbf{x'}) = \sigma^2 \exp \left( - \frac{|| \mathbf{x} - \mathbf{x'}||^2}{2 \ell^2}\right).
\end{align}
The kernel hyperparameter $\sigma$ is a scaling parameter which determines the average distance of the function from the mean, while the hyperparameter $\ell$ is known as a length scale which controls the domain distance at which two points significantly influence one another.

Practically, we are interested in a finite subset of the input domain $X = \{ \mathbf{x}_1, \ldots, \mathbf{x}_n \}$ and sampling function evaluations $\mathbf{y}$ from a multivariate Gaussian marginal distribution defined by mean vector $\mu(X)$ and covariance matrix $k(X, X)$, 
\begin{align} \label{eq:gaussian_process_marginal}
    \mathbf{y} = f(X) \sim \mathcal{N}\left(\mu(X), k(X, X) \right).
\end{align}
$\mu(X) = \mathbf{0}$ (the zero vector), and the covariance function $k$ is the RBF kernel in Eq.~\ref{eq:rbf_kernel}. 

First, we define a domain, $X_d$, and uniformly sample a ``training" set $X_t \subseteq X_d$. We then sample the training targets, $\mathbf{y}_t$ using a GP prior with an RBF kernel. 
\begin{align} \label{eq:y_train_sample}
    \mathbf{y}_t = f(X_t) \sim \mathcal{N}\left(\mathbf{0}, K(X_t, X_t) \right).
\end{align}
Our entire domain $X_d$ serves as ``test" features, allowing us to condition the joint distribution
\begin{align}
    \begin{bmatrix} \mathbf{y}_t \\ \mathbf{y}_d \end{bmatrix} 
    \sim \mathcal{N} \left( 
    \begin{bmatrix} \mu_t \\ \mu_d \end{bmatrix}, 
    \begin{bmatrix} \Sigma_{tt} & \Sigma_{td} \\ \Sigma_{dt} & \Sigma_{dd} \end{bmatrix} 
    \right),
\end{align}
upon the training data $\mathcal{D}= (X_t, \mathbf{y}_t)$. The resulting distribution is given by 
\begin{align} \label{eq:final_posterior}
    p(\mathbf{y}_d | X_d, \mathcal{D})  & = \mathcal{N}(\mu_{f|\mathcal{D}}(X_d), K_{f|\mathcal{D}}(X_d, X_d))\,, \nonumber \\
                                              & = \mathcal{N} ( \mu_{d} + \Sigma_{dt} \Sigma_{tt}^{-1} (\mathbf{y}_t - \mu_{t}), \Sigma_{dd} - \Sigma_{dt} \Sigma_{tt}^{-1} \Sigma_{td})\,.
\end{align}
The distribution in Eq.~\ref{eq:final_posterior} can be sampled efficiently (linear time) using decoupled sampling~\cite{wilson_efficiently_2020}, and is used to generate pairs of cheap and expensive surfaces. All GP code used in this work is implemented using the \textit{Python} package \gpflow~\cite{matthews_gpflow_2017,van_der_wilk_framework_2020}.

Synthetic response surface pairs are binned based on their Spearman coefficient, $r_s$. Bins have a width of 0.25, giving 8 bins on [-1, 1]. The same 20 expensive surfaces, $ \{ \mathbf{y}_{\text{exp}, i} \}_{i=1}^{20}$, are used in each bin while the corresponding cheap surfaces are selected such that their $r_s$ falls into one of the bins. With this setup, a predictive model which has no knowledge of the cheap surfaces should have consistent average performance across all bins.

For input and output dimensionality of 1, the domain is defined as $X_d \in [-5, 5]^{100}$. The lengthscale, $\ell=1.0$ and variance $\sigma^2 = 2.0$ for the RBF kernel in Eq.~\ref{eq:rbf_kernel}. For input dimensionality of 2, $X_d \in [-5, 5]^{10000}$.


Given that the number of training points on the expensive surfaces is often not enough to resolve the entire surface, a cross-validation-like strategy is used for dataset generation, whereby we randomly sample $n_{\text{folds}}$ batches of training points on the expensive surfaces and $n_{\text{folds}}$ batches of training points on the cheap surface and use the remainder of data for validation. The algorithm's average prediction accuracy across all validation folds is reported. For all analytic tests considered in this work, we use $n_{folds} = 20$ The size of the training sets are given in Tab~\ref{tab:analytic_test_scaling}.

\begin{figure}[!ht]%
    \begin{center}
    \includegraphics[width=\textwidth]{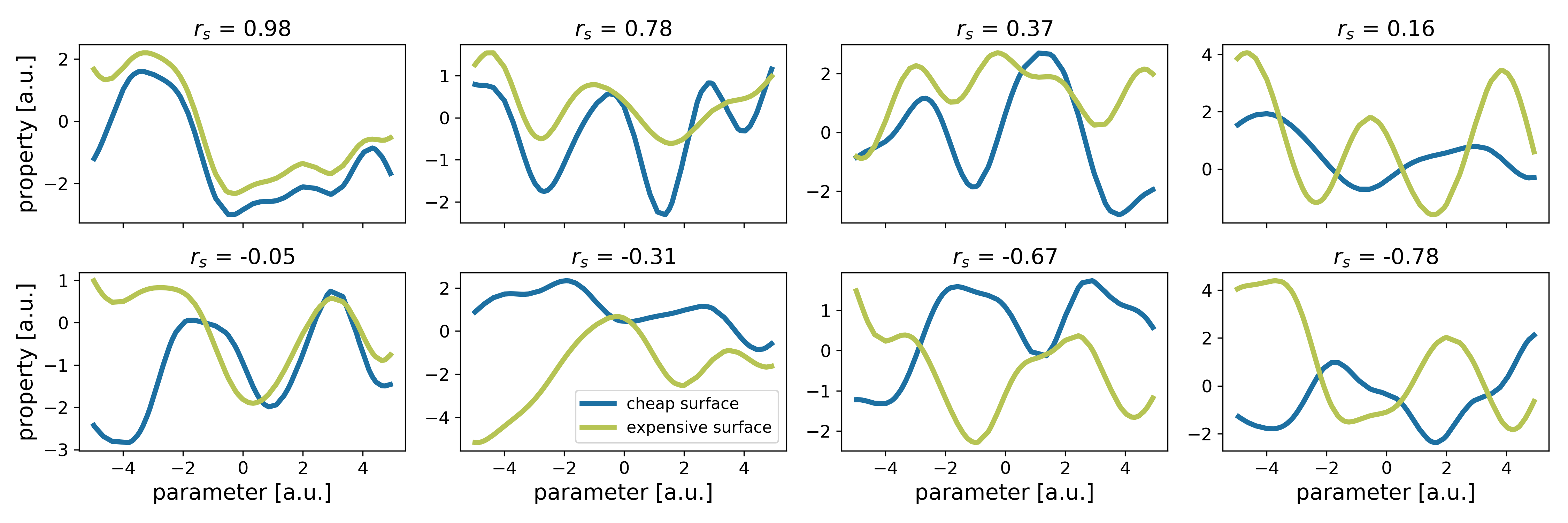}
    \end{center}
    \caption{Synthetic response surface pairs sampled from a GP posterior. This plot contains one pair of surfaces from each of the 8 bins defined by Spearman's rank correlation coefficient between the surfaces. Surfaces range from highly correlated (top-left), to highly anti-correlated (bottom-right).}
    \label{fig:sample_surface_pairs}%
\end{figure}

\begin{center}
    \begin{table} \label{tab:analytic_test_scaling}
        \begin{tabular}{ | l | l | l | l | l | p{5cm} |} 
        \hline
        input dim & output dim & \# total points & \# folds & \# cheap train & \# expensive train  \\ \hline
         1 & 1  & 100 & 20 & 75 & \{2, 3, 5, 10, 20, 50, 75\} \\ \hline 
         2 & 1 & 10000 & 20 & 7500 & \{2, 3, 5, 10, 20, 50, 100, 500, 5000\} \\ 
        \hline
        \end{tabular} 
        \caption{Number of folds, number of cheap training points and number of expensive training points.}
    \end{table}
\end{center}

\subsection{Analytic tests using the Olympus package} \label{subsub:anal_results_olympus}

In high dimensions, sampling response surface pairs with tailored correlations using the GP approach described above becomes costly. Instead, several well-known synthetic surfaces implemented in the \olympus package~\cite{hase_olympus_2020} are selected for high-dimensional tests. 

\begin{figure} [htb]
    \centering
    \includegraphics[width=\textwidth]{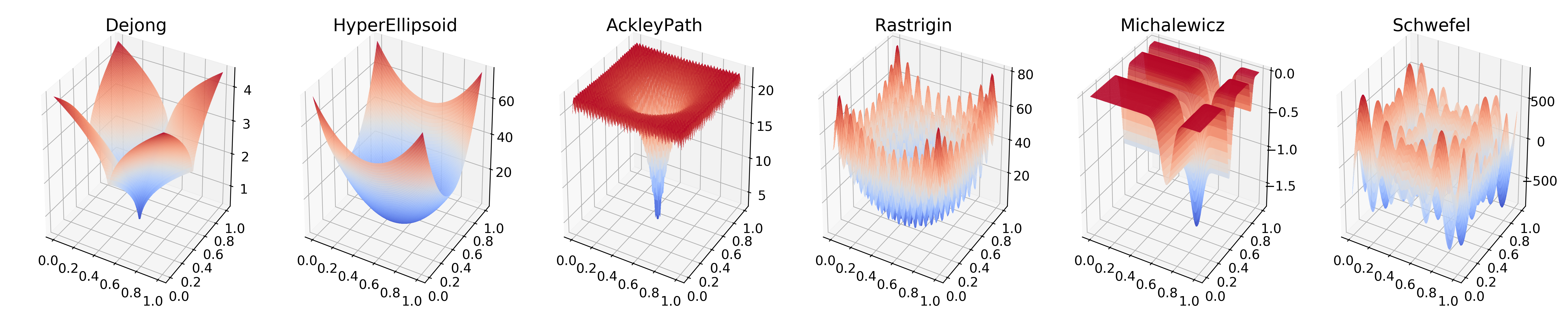}
    \caption{2D synthetic surfaces from the \olympus package considered in this work.}
    \label{fig:olympus_surfaces}
\end{figure}

\begin{table}[!ht] \label{tab:olympus_corrs}
\caption{Spearman rank correlation coefficient between the synthetic surfaces measured on a 2D grid of $1e4$ points.}
\begin{tabular}{l|llllll}
       & Dejong & HyperEllipsoid & AckleyPath & Rastrigin & Michalewicz & Scwefel \\ \hline
Dejong & 1.00   & 0.88           & 0.72      & 0.66        & 0.46  & 0.00      
\end{tabular}
\end{table} 

Namely, the \texttt{Dejong} surface is the expensive surface, while the \texttt{Dejong}, \texttt{HyperEllipsoid},  \texttt{Rastrigin}, \texttt{Michalewicz},  \texttt{Schwefel} and  \texttt{AckleyPath} surfaces are the cheap surfaces. 2D visualizations of these surfaces are given in Fig.~\ref{fig:olympus_surfaces}. The Spearman coefficients between expensive and cheap surfaces are summarized in Table~\ref{tab:olympus_corrs}. The examples include strong (\texttt{Dejong}-\texttt{HyperEllipsoid}), moderate (\texttt{Dejong}-\texttt{AckleyPath}, \texttt{Dejong}-\texttt{Rastrigin}), weak (\texttt{Dejong}-\texttt{Michalewicz}), as well as no correlation (\texttt{Dejong}-\texttt{Schwefel}). 

\begin{figure} [!ht]
    \begin{center}
    \includegraphics[width=\textwidth]{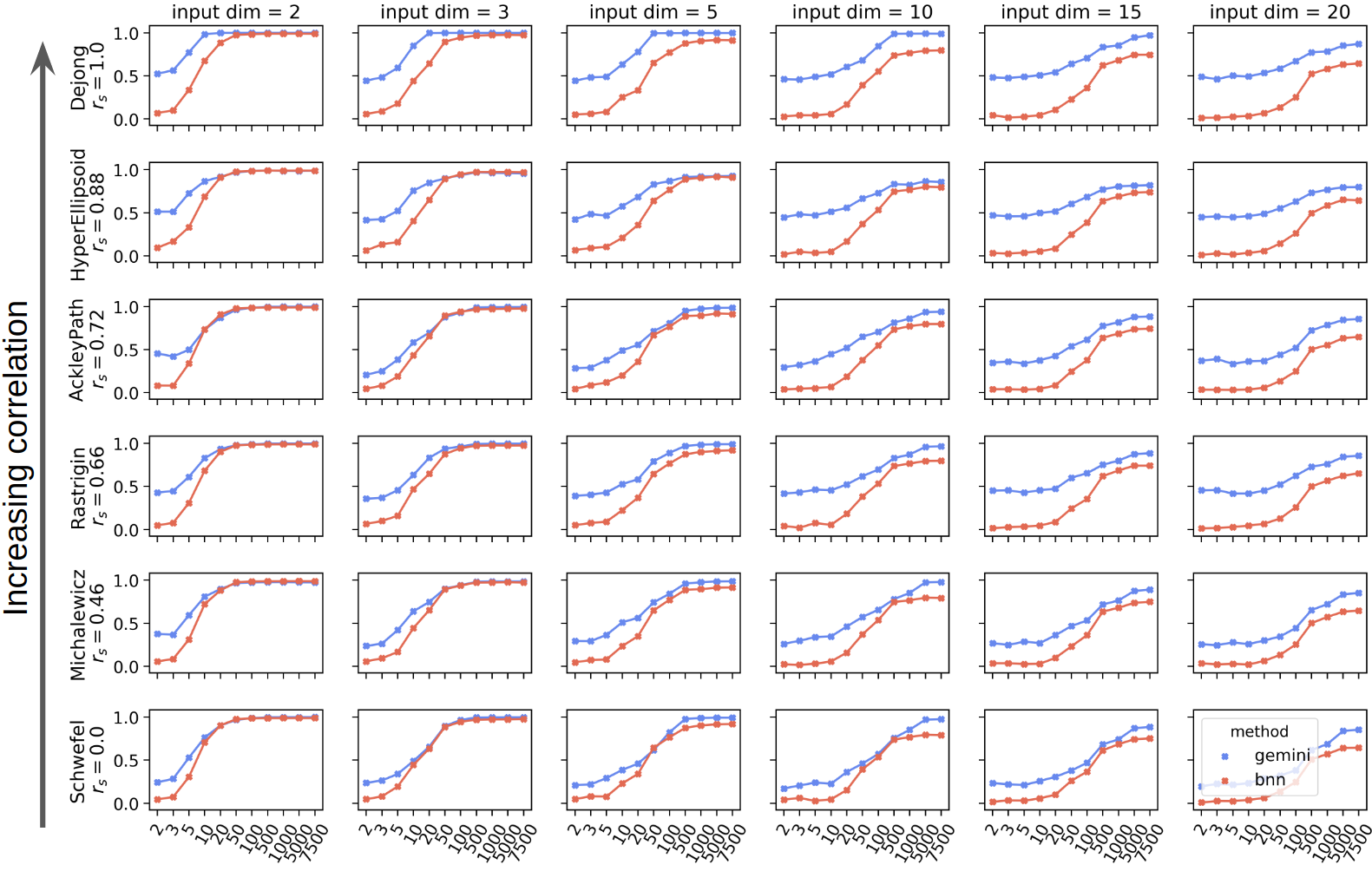}
    \end{center}
    \caption{Learning curves for regression tests using pairs of well-known benchmark surfaces.}
    \label{fig:olympus_surfaces}
\end{figure}

Fig.~\ref{fig:olympus_surfaces} shows learning curves comparing the performance for these tests for input/parameter dimension of 2, 3, 5, 10, 15, and 20. Each subplot row corresponds to a different cheap surface, while \texttt{Dejong} is always used as the expensive surface. The rows are arranged in order of descending correlation (top to bottom) with the expensive surface. In each panel, a learning curve is plotted for \gemini and a vanilla NN (trained only on the expensive points) using Pearson's correlation coefficient as the performance metric. We use a domain of 50000 randomly sampled points, 75\% of which are used as training points on the cheap surface, the rest for validation. The expensive surface training set size is varied along the horizontal axis of the subplots, and the remainder are used for validation. Each point on the subplot is an average performance taken over 10 randomly selected folds. 

We observe the general trend that as correlation increases (ascending a column), \gemini's predictions of the expensive surface become increasingly linearly correlated with the true surface compared to the vanilla NN. We observe that in virtually all the subplots the difference between the performance of \gemini and that of the vanilla NN is greatest is the low-data regime (2-20 points). This observation reinforces the claim that \gemini is an effective tool for regression problems where expensive data is scarce and cheap data is comparatively abundant.

\section{Architecture of the neural networks} \label{subsec:architecture}

All models considered in this work are neural networks implemented using TensorFlow~\cite{abadi_tensorflow_2016}. The dimensionality of the input layer is given by that of the parameter space. All layers in the networks with the exception of the output are connected using either the leaky ReLU or softplus activation functions, the former of which parameterized by a $``$negative slope" of 0.2. The output layer activation depends on the physical meaning of the targets. The reader is referred to SI Sec.~\ref{subsec:hyperparams} for tabulated model hyperparameters. 

We construct a NN with $n$ hidden layers as
\begin{align}
       & \text{batch normalization}  \nonumber \\
\phi_1 & = \text{act hidden}(\mathbf{x} \cdot w_0 + b_0)\,, \\ 
       & \text{batch normalization} \nonumber \\
\phi_2 & = \text{act hidden}(\phi_1 \cdot w_1 + b_1)\,, \\
& \qquad \qquad \vdots \nonumber \\
\phi_n & = \text{act hidden}(\phi_{n-1} \cdot w_{n-1} + b_{n-1})\,, \\ 
       & \text{batch normalization} \nonumber \\
\phi_{\text{out}} & = \text{act out}(\phi_{n} \cdot w_{\text{out}} + b_{\text{out}})\,.
\end{align}
This core NN architecture will be referred to as $\mathcal{F}(\mathbf{x})$ from herein. \gemini is constructed from 3 such NNs. The parameter bias network, $\mathcal{F}_{P}$, the latent network $\mathcal{F}_{L}$, and the target bias network, $\mathcal{F}_{T}$. There are two $``$branches" in \gemini, corresponding to the computational graphs that transform cheap and expensive parameters to predictions (see main text Fig.~\ref{fig:gemini_concept}). For a vector of cheap parameters $\mathbf{x}_{\text{cheap}}$, predictions on the cheap surface are generated by the forward pass
\begin{align} \label{eq:cheap_forward_pass}
\phi_{\text{out}}^{\text{cheap}} & = \mathcal{F}_{L}(\mathbf{x}_{\text{cheap}})\,, \\
\mu_{\text{cheap}}, \sigma_{\text{cheap}} & =  \text{split}(\phi_{\text{out}}^{\text{cheap}})\,.
\end{align}
Where the network output is split to generate a predictive mean and uncertainty. For expensive parameters $\mathbf{x}_{\text{exp}}$, predictions on the expensive surface are generated by the forward pass
\begin{align} \label{eq:exp_forward_pass}
\phi_{\text{out}}^{\text{exp}}  & =  \mathcal{F}_{L}(\mathbf{x}_{\text{exp}} + \mathcal{F}_{P}(\mathbf{x}_{\text{exp}})) +   \mathcal{F}_{T}(\mathcal{F}_{L}(\mathbf{x}_{\text{exp}} + \mathcal{F}_{P}(\mathbf{x}_{\text{exp}})))\,, \\ 
\mu_{\text{exp}}, \sigma_{\text{exp}}  & = \text{split}(\phi_{\text{out}}^{\text{exp}})\,.
\end{align}

We use densely connected deterministic layers in all models in this work. The point estimate weights and biases  $w_i$ and $b_i$ are updated during training. We avoid using computationally more expensive reparameterization estimates (such as in Ref.~\cite{kingma_variational_2015}) by using point estimates on the weights and outputting a mean and standard deviation instead of representing the weights an biases of the network with probability distributions. In this discussion we refer collectively to trainable parameters of each network of \gemini as  $\boldsymbol{\theta}^{\mathcal{F}_P}$, $\boldsymbol{\theta}^{\mathcal{F}_T}$, and $\boldsymbol{\theta}^{\mathcal{F}_L}$ respectively.

To train \gemini, a custom loss function which is a linear combination of the losses of each branch is defined. The variance can never be negative or zero, so we use the logistic function, $S$ plus the addition of a small positive constant to enforce this, i.e. $\sigma_{\text{exp}}=S(\sigma^{\text{net}}_{\text{exp}})+0.1$ and $\sigma_{\text{cheap}}=S(\sigma^{\text{net}}_{\text{cheap}})+0.01$. Using these parameters, we define the normal distributions
\begin{align}
p({y_{\text{exp}}| \mathbf{x}_{\text{exp}}}) & = \mathcal{N} \left( \mu_{\text{exp}}, \sigma_{\text{exp}}\right)\,, \\
p(y_{\text{cheap}}| \mathbf{x}_{\text{cheap}}) & = \mathcal{N} \left( \mu_{\text{cheap}}, \sigma_{\text{cheap}}\right)\,.
\end{align}
Taking the natural log of both $p({y_{\text{exp}}| x_{\text{exp}}})$ and $p(y_{\text{cheap}}| x_{\text{cheap}})$ leads to log likelihood expressions. Ignoring the constant terms and taking the negative of the remaining value, we arrive at expressions that are minimized over all training exmaples, $i$.
\begin{align}
\mathcal{L}_{\text{exp}} = \sum_i \frac{1}{2} \left( \frac{(y_{\text{exp}}^{i} - \mu_{\text{exp}})^2 }{ \sigma_{exp}^2 }   + \log(\sigma_{\text{exp}}^2) \right)\,,
\end{align}
\begin{align}
\mathcal{L}_{\text{cheap}} = \sum_i \frac{1}{2} \left( \frac{(y_{\text{cheap}}^{i} - \mu_{\text{cheap}})^2 }{ \sigma_{cheap}^2 }   + \log(\sigma_{\text{cheap}}^2) \right).
\end{align}

The final loss function for \gemini is as follows, 
\begin{align}
\mathcal{L} = \mathcal{L}_{\text{exp}} + \xi \mathcal{L}_{\text{cheap}} + \lambda_{\text{bias}}\left( || \boldsymbol{\theta}^{\mathcal{F}_P} ||^2  + || \boldsymbol{\theta}^{\mathcal{F}_T} ||^2 \right) + \lambda_{\text{latent}} || \boldsymbol{\theta}^{\mathcal{F}_L} ||^2 .
\end{align}
Where $\xi$ is an additional hyperparameter and the final three terms are L2 regularizations on the parameters of $\mathcal{F}_P$, $\mathcal{F}_T$, and $\mathcal{F}_L$ respectively, with hyperparameters $\lambda_{\text{bias}}$ and $\lambda_{\text{latent}}$. $\mathcal{L}$ is minimized by stochastic gradient descent using the Adam optimizer~\cite{kingma_adam:_2014}.

\section{Gemini hyperparameter search}\label{subsec:hyperparams}

10 pairs of synthetic surfaces are used in a hyperparameter search for \gemini. The surfaces are selected to be diverse in several ways: dimensionality of the parameter space, dimensionality of the target space, ratio of expensive to inexpensive measurements and the correlation of the surfaces that make up a pair (Spearman coefficient). The objective for the hyperparameter search is the average cross-validated performance taken over the 10 pairs of surfaces for a given set of hyperparameters. The hyperparameter search space is generated as a random subset of 500 from the large set of possible architectures. The bounds on the individual hyperparameters are outlined in Table~\ref{tab:gemini_grid_search}. The set of hyperparameters that yielded the best objective value out of all considered are also given in the optimal columns. These hyperparameters ship with the \gemini package and assure the user that the algorithm will perform satisfactorily $``$out-of-the-box" on a diverse set of surfaces.

\setlength{\tabcolsep}{3pt}
\begin{table}[!ht] \label{tab:gemini_grid_search}
\caption{Search ranges and best performing \gemini hyperparameters from the random grid search. Hyperparameters with no entry in the \emph{lower bound} and \emph{upper bound} columns were chosen manually and fixed. For the categorical parameters (act fbias and act tbias), only softplus and leaky ReLU were used in the search. $^{\dagger}$ $P$ is the dimensionality of the input parameters. }
\centering
\begin{tabular}{r | c | c | c}
\toprule
\textbf{Hyperparameter} & \textbf{lower bound} & \textbf{upper bound} & \textbf{optimal} \\ 
\hline
batch size              & 50            & 100            & 50           \\
learning rate           & 5e-5          & 3e-4            & 0.000272         \\
act fbias               & leaky relu   & softplus        & softplus       \\
act fbias out           &            &               & linear      \\
act latent              &            &               & leaky relu   \\
act latent out          &            &               & linear             \\
act tbias               & leaky relu & softplus     & softplus       \\
act tbias out           &            &              & linear           \\
depth fbias             &           &               & 1              \\
depth latent            &           &              & 3            \\
depth tbias             &           &              & 1        \\
hidden fbias            &           &               & $P^{\dagger}$     \\
hidden latent           &           &               & 96           \\
hidden tbias            & 1         & 3              & 3             \\
coeff both $(\xi)$             & 0.5       & 3             & 0.5 \\
reg latent $(\lambda_{\text{latent}})$             &           &               &  1e-3\\
reg bias   $(\lambda_{\text{bias}})$             & 2e-1      & 1e-3               &  0.0894      \\
\bottomrule
\end{tabular}
\end{table}

\section{Discussion of the qualitative effect of parameters $\lambda$ and $\rho$ on the acquisition function} \label{subsec:lambda_rho_acq}

\begin{figure}[!ht]%
    \begin{center}
    \includegraphics[width=0.7\textwidth]{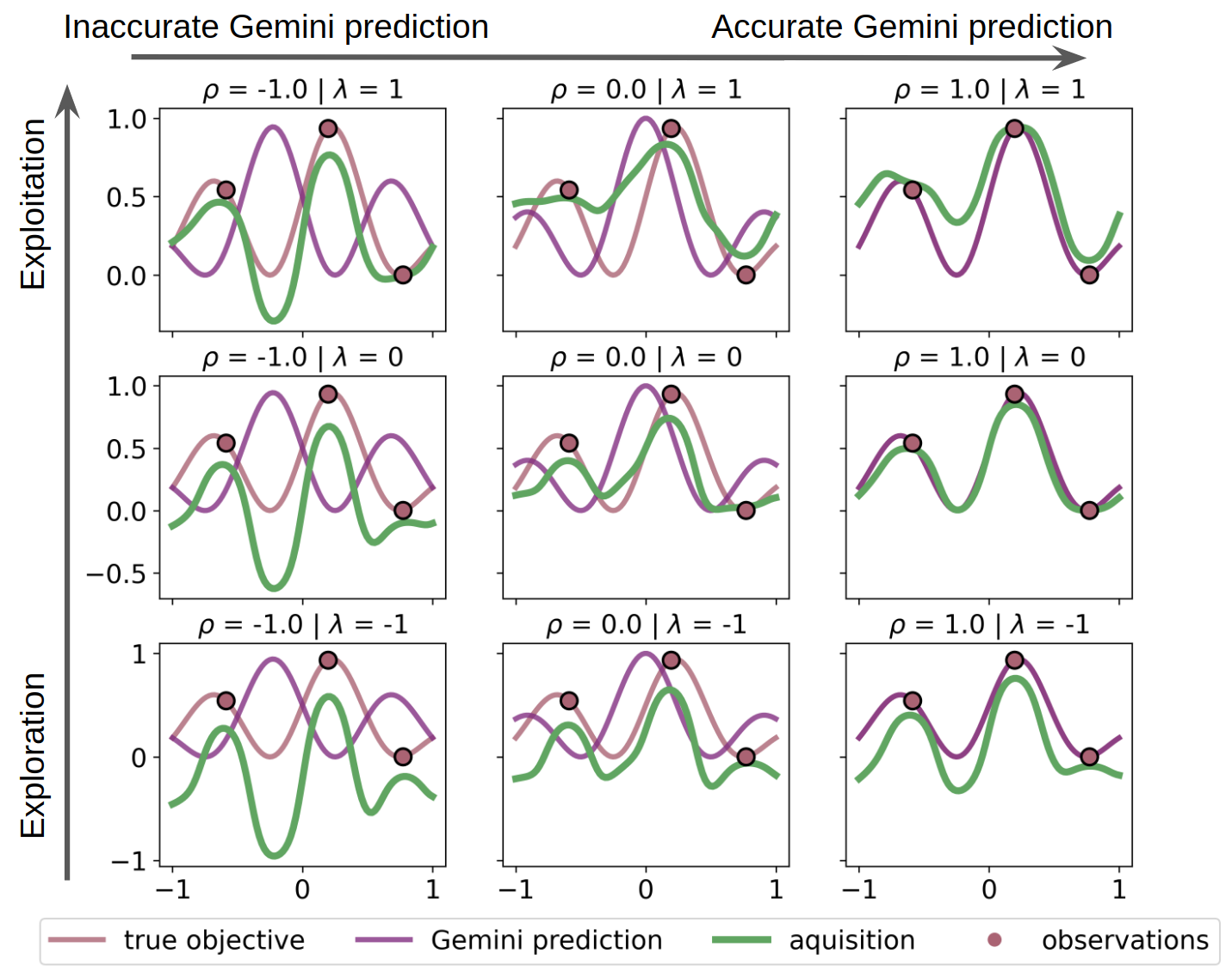}
    \end{center}
    \caption{Depiction of the effect of varying the parameters $\lambda$ and $\rho$ in the \gemini + \phoenics acquisition function $\alpha_{g}(\mathbf{x})$ (Eq.~\ref{eq:gemini_acquisition_function} in the main text). $\lambda$ is a hyperparameter that takes values in $[-1, 1]$. The algorithm is fully biased toward exploration (exploitation) when $\lambda$ is -1 (1). $\rho$ is the Pearson correlation coefficient between the observed expensive surface measurements and the \gemini predictions.}
    \label{fig:lambda_rho_param}
\end{figure}

In the acquisition function (main text Eq.~\ref{eq:gemini_acquisition_function}), $\alpha_{g}(\mathbf{x})$ uses the prediction of \gemini in order to determine subsequent proposals. If \gemini's prediction is uncorrelated ($\rho \approx 0$) with the current observations, the acquisition function diverts back to the pure \phoenics acquisition. When $\rho$ is close to 1, $\alpha_{g}(\mathbf{x})$ will be heavily influenced by \gemini's prediction (right most column of Fig.~\ref{fig:lambda_rho_param}). The acquisition will also be heavily influenced by the \gemini predictions if $\rho$ is close to $-1$ (left most column in Fig.~\ref{fig:lambda_rho_param}), but in an inverse way, i.e. regions in which \gemini predicts to be promising will have a larger $\alpha_g(\mathbf{x})$ and vice-versa. The form of Eq.~\ref{eq:gemini_acquisition_function} is also appealing because it still only depends on one hyperparmeter: $\lambda \in [-1, 1]$, which smoothly biases its behaviour between exploration and exploitation. The effect of $\lambda$ on the shape of $\alpha_{g}(\mathbf{x})$ is depicted in the subplot rows of Fig.~\ref{fig:lambda_rho_param}.

\newpage
\section{Details of perovskite regression experiments} \label{subsec:perovskite_details}

\begin{figure}[!ht]%
    \begin{center}
    \includegraphics[width=0.4\textwidth]{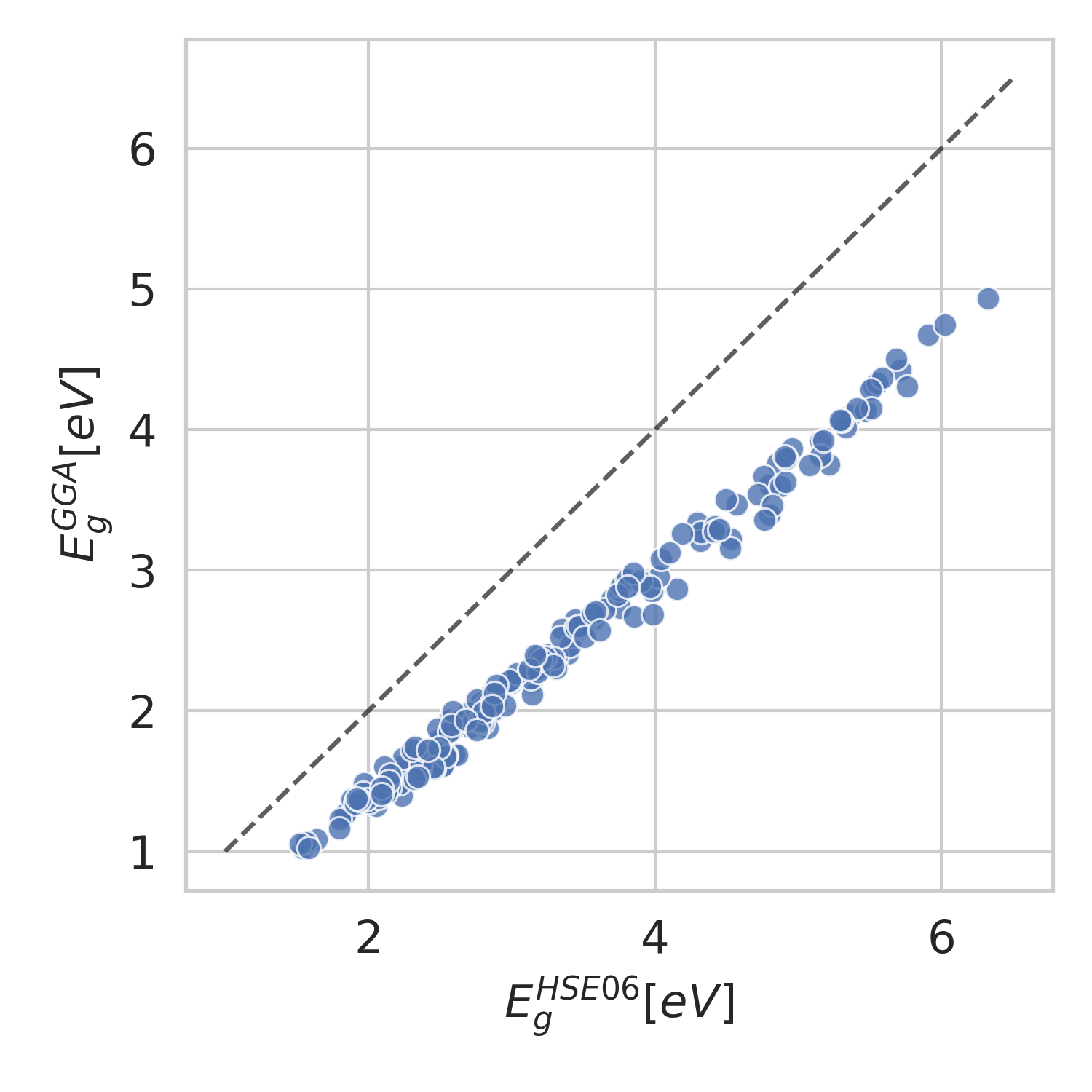}
    \end{center}
    \caption{Parity plot between the calculated bandgaps computed at the GGA and HSE06 levels of theory. The GGA level of theory systematically underestimates the bandgap. The values are nearly perfectly monotonically related with $r_s=0.99$.}
    \label{fig:perovskite_agreement}%
\end{figure}

Fig.~\ref{fig:perovskite_agreement} shows the relation between the bandgaps calculated at the two levels of theory reported in the HOIP dataset~\cite{kim_hybrid_2017}.

Below are listed the physicochemical descriptors and their values which are combined into a 14 element vector to describe each of the 192 pervoskite materials in the HOIP dataset. Inorganic components are characterized by their electron affinity, ionization energy, mass, and electronegativy. The organic anions are characterized by their HOMO and LUMO energies, dipole moment, radius of gyration and total mass. The authors of Ref.~\cite{hase_gryffin_2020} report a significant degree of correlation between many of these descriptors and the calculated bandgaps in the HOIP dataset. The reader is directed to Ref.~\cite{hase_gryffin_2020} SI Sec. S.4.B. for additional details.

\section{General details of the \gemini + \phoenics optimizer} \label{sec:general_opt_details}

In a closed-loop workflow, \gemini is trained only when the number of expensive observations $\geq 2$, because the Pearson coefficient is undefined with 1 point. At each subsequent iteration, \gemini is trained using a cross-validation scheme. At each iteration the dataset of observations is split into 3 cross-validation folds. \gemini is trained and makes prediction on each of the folds and the average Pearson coefficient value over the folds is taken as $\rho$ in Eq.~\ref{eq:gemini_acquisition_function}, i.e.
\begin{align}
\rho_{\text{cv}} = \frac{1}{3} \sum_{i=1}^3 \rho \left( g(\mathbf{x}_i), f(\mathbf{x}_i) \right)\,.
\end{align}
Where $g(\mathbf{x}_i)$ is \gemini's prediction of the expensive targets of the $i^{\text{th}}$ validation fold and $f(\mathbf{x}_i)$ are the expensive observations themselves. $\rho\left(g(\mathbf{x}_i), f(\mathbf{x}_i)\right)$ is given by
\begin{align} \label{eq:pearson}
\rho\left(g(\mathbf{x}_i), f(\mathbf{x}_i)\right) = \frac{\mathbb{E} \left[ \left(g_{i,j} - \mu^g_{i}\right) \left( f_{i, j} - \mu^{f}_i\right) \right]}{ \sigma^{g}_i \sigma^{f}_i}\,.
\end{align}
Where $\mu^{g}_i$ and $\sigma^{g}_i$ are the mean and standard deviation of the predictions on the $i^{\text{th}}$ fold. $\mu^{f}_i$ and $\sigma^{f}_i$ are the mean and standard deviation of the true objective values of the $i^{\text{th}}$ fold.
 
\section{Details of optimization of electrocatalytic activity of high-dimensional composition spaces for the oxygen evolution reaction}
\label{subsec:details_opt_cat_oer}

The hyperparameters of the BNN used to emulate the discrete OER catalyst experiments are given in Table~\ref{tab:emulator_hyparams_cat_oer}. The hyperparameters were determined manually. Networks are trained for a maximum of 30000 epochs, or until a predefined early stopping criteria is met. Networks use the densely connected reparameterization layers implemented in TensorFlow Probability~\cite{abadi_tensorflow_2016} and the ELBO loss function is minimized using the Adam optimizer.  Once trained, the resulting weights and biases of the networks are saved and are made accessible for the optimizer to query with new parameters at each iteration (experiments are emulated as a forward pass of the network using the proposed parameters as input). The elemental compositions of the OER catalysts are transformed as they are passed between the BNN emulator/\gemini and \phoenics. Specifically, optimization takes place on the 5-dimensional hypercube, while the BNN emulators and \gemini receive a 6-simplex as features. This is motivated by the actual parameters lying on the standard 6-simplex: $\Delta^6 = \left \{  (t_0, t_1, t_2, t_3, t_4, t_5) \in \mathbb{R}^6 | \sum_{i=0}^{6} t_i = 1, t_i \geq 0 \forall i \right  \}$.

\setlength{\tabcolsep}{3pt}
\begin{table*}[!ht] \label{tab:emulator_hyparams_cat_oer}
\caption{Hyperparameters of BNN emulators for the OER electrocatalytic activity optimization experiments.}
\centering
  \begin{tabular}{  c | c  }
    \toprule
      \textbf{ Hyperparameter} & \textbf{Value}  \\
    \hline
    \textbf{batch size}              & 100        \\
    \textbf{hidden depth}            & 4          \\
    \textbf{neurons per layer}       & 96         \\
    \textbf{learning rate}           & 3e-4       \\
    \textbf{regularization}          & 1e-4       \\
    \textbf{hidden activation}       & leaky ReLU  \\
    \textbf{out activation}          & ReLU        \\
    \bottomrule
  \end{tabular}
\end{table*}

\begin{figure} [!ht]
    \begin{center}
    \includegraphics[width=\textwidth]{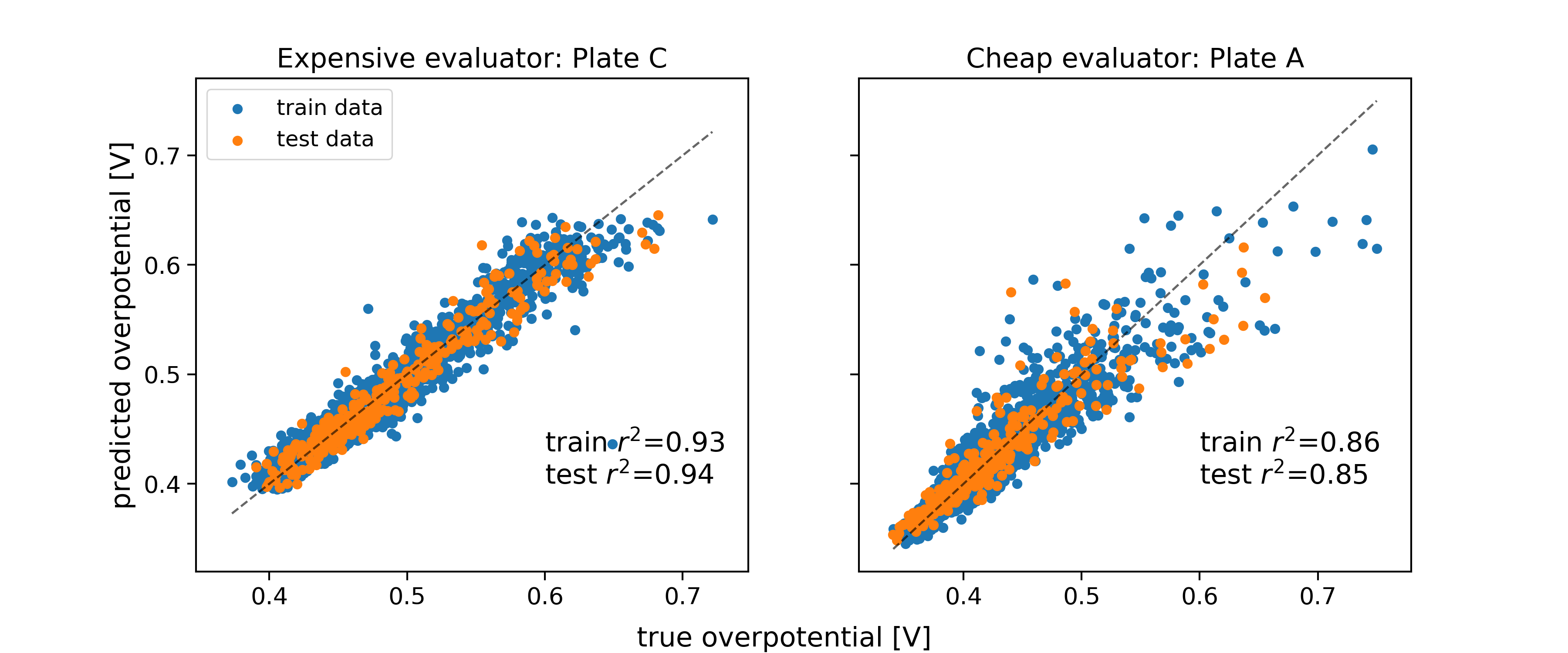}
    \end{center}
    \caption{Prediction accuracies of the BNN experiment emulators. Data is split randomly into 85/15 train/test split.}
    \label{fig:emulated_surfaces.png}
\end{figure}

\begin{figure} [!ht]
    \begin{center}
    \includegraphics[width=\textwidth]{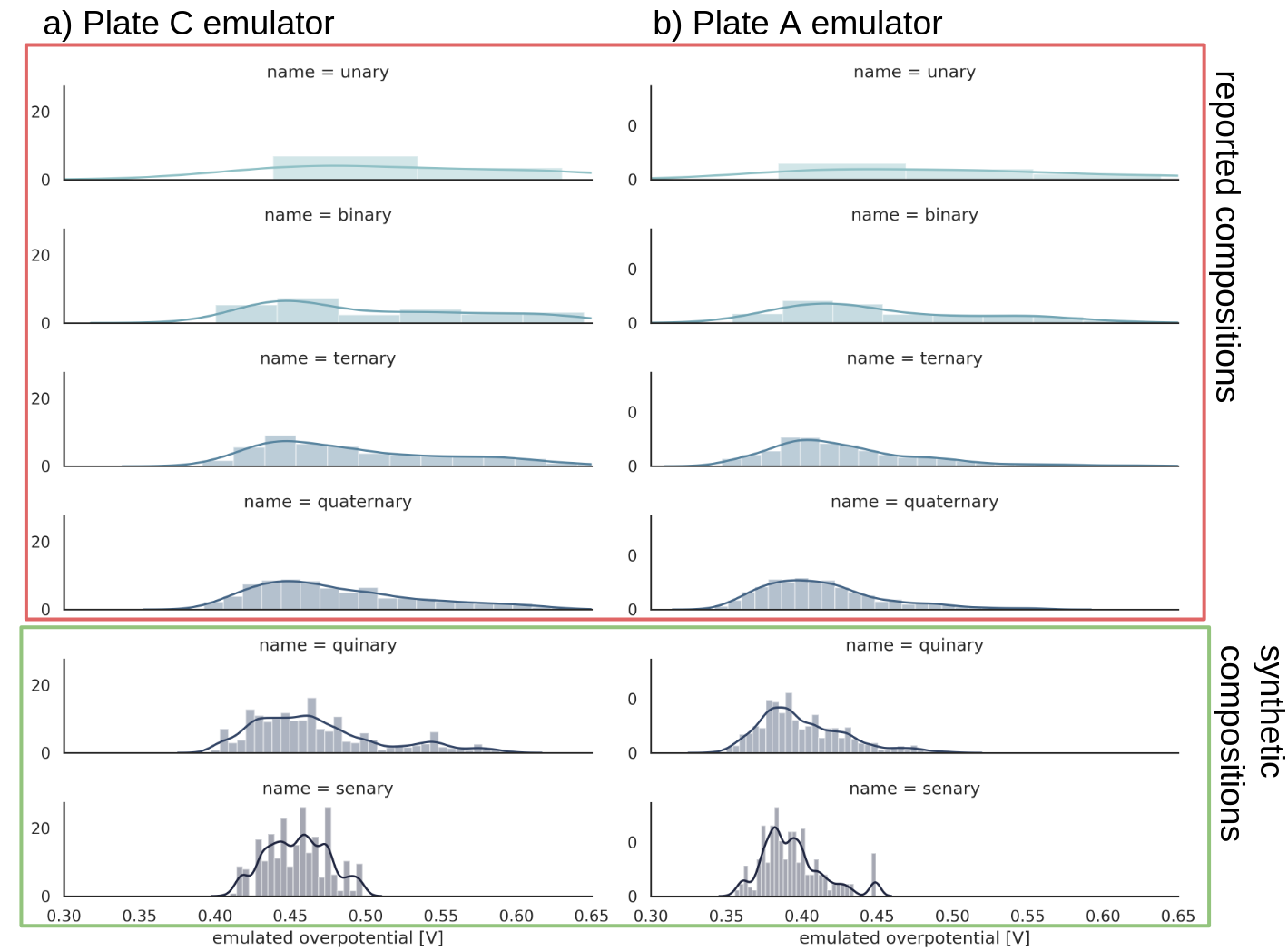}
    \end{center}
    \caption{Comparison of the distributions of catalytic overpotentials predicted by the Plate C and A emulators. Unary, binary, ternary and quinary compositions (red box) are measured in Ref.~\cite{suram_benchmarking_2020}. Quinary and senary compositions (green box) are generated in this work and passed through the emulator.}
    \label{fig:cat_oer_emulator_comparison.png}%
\end{figure}

During the optimization experiments, optimization strategies traverse the entire standard 6-simplex of catalyst compositions. Ref.~\cite{suram_benchmarking_2020} reports only unary, binary, ternary and quaternary compositions from 6 element sets with 10 at\% intervals. For the compositions whose overpotentials are not reported in the original dataset, i.e. the quinary and senary compositions, we would like to know if our probabilistic emulator predicts \emph{reasonable} values for overpotentials with respect to the measured values. We do not claim that these synthetic values are quantitatively accurate compared to experiment. Overpotentials for all possible quinary (5040 points) and senary (3600 points) compositions at 10 at\% intervals are generated using our trained BNN emulator and plotted as distributions in Fig.~\ref{fig:cat_oer_emulator_comparison.png}. The red box surrounds the compositions which are actually measured in the original dataset while the green box surrounds the synthetic compositions. We observe that the distributions of the synthetic compositions are consistent with the ranges of the reported compositions. Therefore we deem these synthetic values reasonable enough for use in our optimization experiments.